\documentclass[preprint,12pt]{elsarticle}
\usepackage[utf8]{inputenc}
\usepackage{hyperref}
\usepackage{colortbl}
\usepackage{pstricks}
\usepackage{pst-all}
\usepackage{tikz,pgfplots}
\usepackage{tabularx}
\usepackage[english]{babel}
\usepackage[babel=true]{csquotes}
\usetikzlibrary{arrows,automata}
\usepackage[margin=2cm]{geometry}
\usepackage{fancyhdr} 
\usepackage{graphics} 
\usepackage{graphicx} 
\usepackage{pstricks,pst-node} 
\usepackage{tikz} 
\usepackage{wrapfig}
\usepackage{animate}
\usepackage{subfig}
\usepackage{algorithm,algorithmic}
\usepackage{comment}
\usepackage{graphicx}
\usepackage{multimedia}
\usepackage{animate}
\usepackage{hyperref}
\usepackage{amsmath} 
\usepackage{ulem}
\usepackage{amssymb}
\pgfplotsset{compat=1.18} 

\begin{document}

\makeatletter
\def\ps@pprintTitle{%
 \let\@oddhead\@empty
 \let\@evenhead\@empty
 \let\@oddfoot\@empty
 \let\@evenfoot\@empty
}
\makeatother

\begin{frontmatter}

\title{Optimizing UAV Trajectories via a Simplified Close Enough TSP Approach}

\author{Hiba Bederina\\
hiba.bederina@gmail.com}

\begin{abstract}
This article explores an approach to addressing the Close Enough Traveling Salesman Problem (CETSP). The objective is to streamline the mathematical formulation by introducing reformulations that approximate the Euclidean distances and simplify the objective function. Additionally, the use of convex sets in the constraint design offers computational benefits. The proposed methodology is empirically validated on real-world CETSP instances, with the aid of computational strategies such as a fragmented CPLEX-based approach. Results demonstrate its effectiveness in managing computational resources without compromising solution quality. Furthermore, the article analyzes the behavior of the proposed mathematical formulations, providing comprehensive insights into their performance.  
\end{abstract}

\begin{keyword}
Combinatorial Optimization \sep Routing Problems \sep CETSP \sep Unmanned Aerial Vehicle.
\end{keyword}

\end{frontmatter}

\section{Introduction}
This article explores the gathering of data of buried sensor nodes distributed in the environment by means of an Unmanned Aerial Vehicle (drone). The sensor nodes are scattered across a diverse field area, encompassing various elements such as crops, permanent meadows, forests, and buildings
. They emit information from a radio module having a circular communication range of several tens of meters. The drone’s objective is to efficiently gather information from all sensors within a limited time-frame.

The theoretical alignment of this scenario corresponds to the Close Enough Traveling Salesman Problem (CETSP). In this context, the drone serves as a mobile data collector, and each sensor’s encompassing compact region in the plane is defined as its neighborhood set, referred to as the emission area. Consequently, the challenge revolves around identifying optimal locations within these circular zones and determining the most efficient sequence for the drone to visit each sensor, minimizing the overall distance costs.

The Close Enough Traveling Salesman Problem (CETSP) has garnered considerable attention in recent decades due to its practical applications, notably in fields such as the data collection from Unmanned
Aerial Vehicle (UAV) \cite{cariou2023evolutionary}, Automated Meter Reading (AMR) \cite{hou2024natural}, and Radio Frequency Identification (RFID) technologies \cite{patelradio}.

Specifically, the CETSP resides within the domain of combinatorial problems. Solving CETSP involves a combination of two optimization approaches.
Firstly, there's a mixed-integer optimization aspect aimed at determining the most efficient route for the drone. This entails connecting all sensors while originating and concluding at the depot, ultimately seeking the shortest overall route.
Secondly, a continuous optimization facet focuses on finding optimal coordinates within the 
communication range of the sensor nodes. Each sensor's 
communication range is represented as a circular disk centered at the sensor's location with a specific radius. This continuous optimization process aims to improve the travel cost by identifying the most effective coordinates within these 
communication ranges for each sensor.

The Traveling Salesman Problem (TSP) \cite{diaz2024mathematical} represents a specific instance of the Close Enough Traveling Salesman Problem (CETSP) when all circular disks 
are reduced to 
a point. This implies that solving the CETSP is at least as complex as solving the TSP itself.

In the CETSP, the tour must navigate through at least one point within each customer's service region (a sensor node is considered as a customer to be visited by the drone). When a tour traverses an area where several disks overlap, all customers associated with those overlapping disks are considered as served. This region where multiple disks intersect is termed a Steiner zone within the context of the problem \cite{csenturk2020steiner}.

Absolutely, the 
CETSP 
is classified as an NP-Hard problem \cite{Korte2008}, and its complexity amplifies as the number of sensors increases. Beyond the inherent complexity associated with being NP-hard, there's an additional challenge in determining representative points within the Euclidean space. This further complicates the computation of the shortest distances within the objective function and adds complexity to verifying whether these points lie within the corresponding disks as required by the constraints of the problem.

\begin{figure}[ht]
\centering
     \includegraphics[scale=0.6]{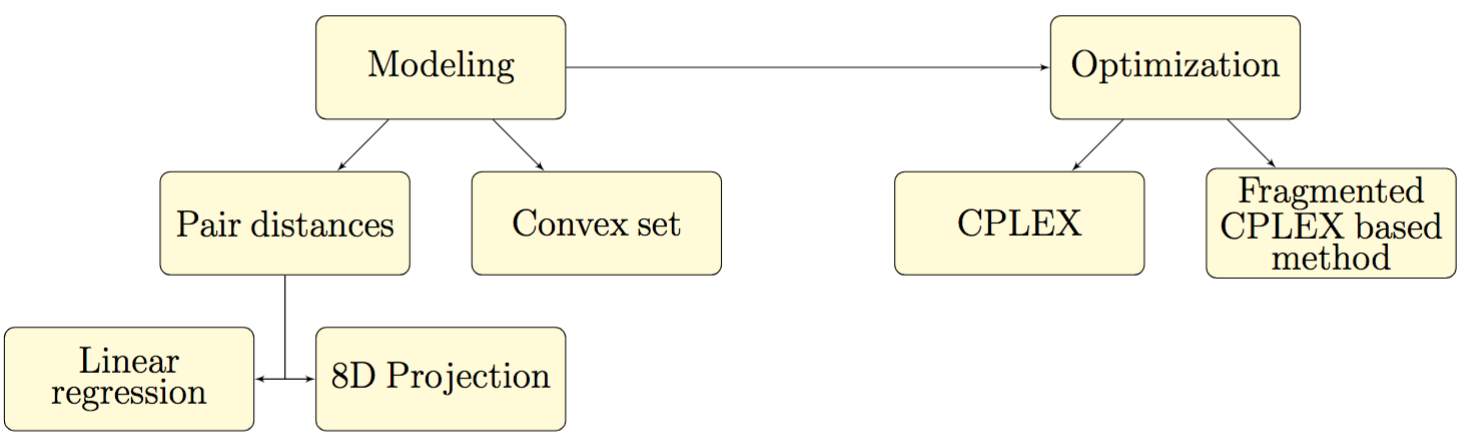}
     \caption{Design and problem solution overview}
     \label{ovv}
\end{figure}

The resolution of the CETSP (see Figure \ref{ovv}), like any optimization problem, unfolds in two distinct phases: modeling and optimization. The modeling phase entails the mathematical formulation of the problem, while optimization involves solving the established model. This paper's contributions focus on streamlining the mathematical formulation by leveraging convex sets to linearize problem constraints. Additionally, the Euclidean distance measure in the objective function is replaced with the Manhattan distance, transforming a cubic equation into a more manageable quadratic one with absolute values and guiding the drone within convex sets, rather than solely along their borders. Furthermore, linear regression enhances the approximation of the Euclidean distance measure, while 8D Projection refines and improves the obtained solutions.

The resulting mathematical model is optimized through two approaches. Firstly, the overall model is solved using CPLEX. Additionally, a Fragmented CPLEX-based method is proposed, addressing the same mathematical problem in manageable chunks and effectively managing intersections between emission fields without introducing randomness.

The remainder of the paper is organized as follows. After the introduction, Section 2 presents the state of the art of CETSP problem. In Section 3, a description of the non linear mathematical formulation of the problem is presented and the concerns related to its resolution are discussed. Then, in Section 4, we give some alternatives in an attempt to linearize the constraints and propose some changes in the objective function. Section 5 shows the results and compares of the proposed models. At the end, Section 6 presents our conclusions and future research directions.

\section{Background}
The Close Enough Traveling Salesman Problem (CETSP) is an extension of the classic TSP problem and is viewed as a specific instance  within three other TSP-related problems, as stated in \cite{mennell2009heuristics}. These include the Traveling Salesman Problem with Neighborhoods (TSPN) \cite{pacheco2023exponential}, the Generalized Traveling Salesman Problem \cite{pop2023comprehensive}, and the Covering Tour Problem \cite{ota2023flow}.

Three primary steps were delineated in \cite{gulczynski2006close} for addressing the CETSP problem with identical radii. Initially, it involves defining super nodes, particularly beneficial for service regions
, ensuring each point falls within at least one disc. Subsequently, a basic TSP problem is tackled based on the locations of these super nodes to determine a pseudo-optimal path. Finally, an optimized function is employed to refine the locations of points to be visited within the area, ensuring the solution's viability.

Then, the focus was on addressing the Optimal Robot Routing Problem (ORRP) \cite{yuan2007optimal} by treating it as a Traveling Salesman Problem with Neighborhoods (TSPN) involving disjoint discs of a specified radius. The approach introduced a two-phase algorithm: initially breaking down the TSPN into a combinatorial problem, aiming to solve the TSP based on the original coordinates. This was followed by a continuous optimization phase, utilizing an Evolutionary Algorithm to determine the hitting points effectively.

Building on the findings of \cite{gulczynski2006close}, the studies by \cite{mennell2009heuristics} and \cite{mennell2011steiner} introduced the Steiner Zone heuristic (SZH), a three-phased approach. Initially, it involves pinpointing intersections among disks, termed Steiner Zones, which enable routing to cover multiple disks concurrently. Second, each identified Steiner Zone is represented by a point (referred to as a super-node or hitting point). Thirdly, a TSP tour is established based on these representative points. Subsequently, the previously determined TSP sequence undergoes enhancement using a TPP (Touring a sequence of Polygon Problem) by relocating these representative points.

Addressing the continuous optimization challenge in CETSP often involves employing discretization techniques to transform continuous covering neighborhoods into clusters. 
Two partitioning schemes for discretizing each neighborhood were introduced in \cite{behdani2014integer} : grid-based and arc-based partitioning. 
Subsequently, 
optimal solutions using an exact Branch and Bound algorithm were achieved in \cite{coutinho2016branch}.

In \cite{carrabs2017novel} 
a novel discretization approach surpassing previous methods, incorporating both Perimetric and Internal Point discretization schemes was proposed. They transformed CETSP into a Generalized TSP (GTSP) using an arc discretization method followed by a graph reduction.

In \cite{yang2018double}, the approach involved decomposing the Traveling Salesman Problem with Neighborhoods (TSPN) into two distinct problems. The first part focuses on a combinatorial challenge, seeking the path sequence for the TSP across the original vertices by employing a Genetic algorithm. Meanwhile, the second aspect tackles a continuous optimization problem using Particle Swarm Optimization (PSO) coupled with Boundary Encoding (BE). This method implements search space reduction techniques to effectively determine the hitting points.

\cite{wang2019steiner} devised the Steiner Zone Variable Neighborhood Search (SZVNS) heuristic, addressing CETSP in three phases: problem size reduction, Steiner zone identification, and Variable Neighborhood Search. Wheras in 
\cite{semami2019close}, a three-phase heuristic centered around Nearest Neighborhood Search (NNS) was introduced to tackle CETSP incorporating time windows.
 
\cite{fanta2021close} proposed two approaches, combining solver and heuristics, to tackle the CETSP problem. GLNS-CETSP, based on a solver and a heuristic, leading to a faster approach but 
with less optimal results, extended to consider obstacles.

Some works considered evolutionary algorithms to solve CETSP. 
\cite{cariou2023evolutionary} introduced a straightforward genetic algorithm utilizing geometry-based operators, while 
\cite{di2023generalized} proposed a heuristic algorithm and a genetic algorithm (GA) to optimize the routing for the Generalized Close Enough Traveling Salesman Problem (GCETSP).

Furthermore, 
\cite{qian2023solving} introduced the Close Enough Orienteering Problem (CEOP), extending CETSP by introducing variable prize attributes and non-uniform cost considerations for prize collection.

Lately, machine learning techniques start to gain attention for combinatorial optimization. 
\cite{sinha2021estimating} employed regression models to determine optimal coefficient values for an equation correlating parameters, aiding in identifying accurate routing length estimation.

\section{Mathematical formulations}
\begin{figure}[ht]
\centering
    \includegraphics[scale=0.33]{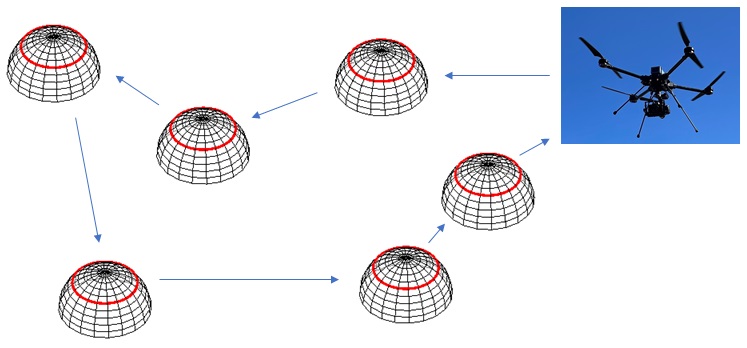}
    \caption{$n = 6, m = 1,\textcolor{blue}{T}, min\ cost(\textcolor{blue}{T})$.}
    \label{Cycle}
\end{figure}

Assuming that the CETSP can be represented as a graph as illustrated in Figure \ref{Cycle}, where the $n$ communicating sensors are defined by the set of vertices $\mathcal{S} = \{0,1,..,i,j,..,n\}$
. A vertex $i$ is located in a Cartesian space at the coordinates $(m_i,n_i)$. 
These coordinates represent the center of a circle of radius $r_i$ delimiting its  communication range. Vertex $0$ represents the departure and arrival station of the drone, 
its radius is fixed to zero in the model.

The paths relating each two sensors are defined by a set of arcs {$E = \{(i,j)/ i,j \in \mathcal{V}\}$}. Each arc is labeled by a cost $d_{ij}$ determining the distance between the vertices $i$ and $j$. $m$ is the number of drones, $T$ is the set of routings (cycles) traveled by the drones starting and ending in the station $0$. In our case, only one drone is used $m=1$. 

$x_{ij}$ is a boolean variable equal to $1$ if the arc $(i,j)$ is travelled by the drone and $0$ otherwise. Some integer variables $u_i$ are needed for each sensor $i$. They represent their order number in the routing list and help to avoid having sub-cycles in the routing. Then, these variables are used in the constraints below in order to find a cycle exactly similar to the Travel Salesman Problem by deciding which arcs are visited and in which order.

In the CETSP, the drone doesn't have necessarily to pass by the centre point of the circle, instead  it will just have to pass by any hitting point $c_i$ inside of the circle. This hitting point will have the coordinates $(cx_i,cy_i)$.

The solution to the problem consists at adding some other constraints to find the well-suited value for each of the previously mentioned variables to have the lowest possible travel cost. In this case the distance matrix value is no longer an input, but that needs to be carried out with the Euclidean distance using the new points (hitting points). The model formulation for the CETSP is defined as follows:

    \begin{align}
    &\min \qquad && \sum_{(i,j)\in E} x_{ij} \cdot [ (cx_i - cx_j)^2  + (cy_i - cy_j)^2 ]  \label{OBJSQ}\\ 
    & \text{s.t.} \qquad && \nonumber \\
    &&&  (cx_i - m_i)^2  + (cy_i - n_i)^2  \leq r_i^2  \qquad \qquad \forall i \in \mathcal{V} \label{MP8}\\
    &&& \sum_{j=0}^n x_{ij} = 1 \qquad \qquad \forall i \in \mathcal{V} \label{MP3} \\
    &&& \sum_{j=0}^n x_{ji} = 1 \qquad \qquad \forall i \in \mathcal{V} && \label{MP4} \\
    &&& x_{ij} + x_{ji} \leq 1 \qquad \qquad  \forall (i,j) \in E && \label{MP5} \\
    &&&  u_{i} - u_j + n \cdot x_{ij} \leq n - 1 \qquad \qquad  \forall (i,j) \in E && \label{MP} \\
    &&& x_{ij} \in \{0,1\}, u_{i} \in \mathbb{N} / u_0=0 &&  \label{MP7}
\end{align}

The objective function (\ref{OBJSQ}) minimizes the Euclidean distance between any sensors $i$ and $j$ where  $x_{ij}=1$, which means belonging to the cycle. The square root is removed from the formulation as it does not affect the optimization. Constraint \ref{MP8} delimits the region where the new point $(cx_i,cy_i)$ is located by ensuring that the distance, separating it from the centre, is less than or equal to the radius of sensor region. Constraint \ref{MP3} ensures that the number of arcs $x_{ij}$ getting out from sensor region $i$ sums to one as only one arc (equal to $1$, and the other equals to zero) is travelled one time by the drone. On the other hand, constraint \ref{MP4} ensures that the sum of arcs $x_{ji}$ getting in the sensor region is equal to one. These two constraints allow the drone to visit the drone only one time.

In our case we considered a directed graph representation for the problem. Therefore, constraint \ref{MP5} requires the drone to cross an arc in one direction. Finally, constraint \ref{MP} helps in finding one cycle and deletes sub-cycles by assigning an order number $u_i$ to each sensor from $1$ to $n$. 
At the end, the variable domains are defined where $x_{ij}$ for each arc $(i,j)$ is binary and $u_i$ is a positive integer for each sensor $i$ (equation (7)).

This formulation for CETSP is quadratic, as there are a quadratic objective function and a quadratic constraint. These quadratic terms will increase the computational time with the increase of the number of sensors. Besides, an automatic solver may not be able to terminate the calculations within a reasonable time and return at least a solution close to optimal.

For each sensor $i$, we propose introducing a convex set that represents the area in which the sensor transmits and where the drone can collect its data. This formulation allows us to handle specific cases—such as squares or hexagons—as convex regions.

\subsection{CETSP mathematical formulation : Small square}
\begin{figure}[ht]
\centering
    \includegraphics[scale=0.25]{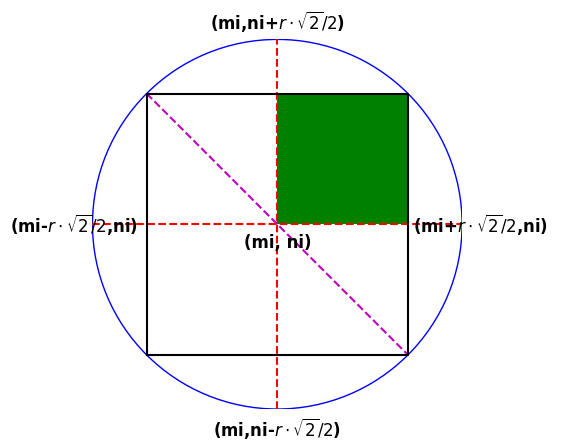}
    \caption{Inscribing a square in a circle.}
    \label{squareincircle}
\end{figure}

This comes down to studying the case where emission areas for each sensor $i$ can be modeled by a square (simple form), that means inscribe a square inside each circle as illustrated in Figure \ref{squareincircle} to linearize equation \ref{MP7}.

This form is so simple to consider; it just needs to delimit the domain values of $cx_i$ and $cy_i$ for each sensor depending on its centre and radius, without the need to add complicated constraints. For this, some geometry will be considered. Let define $l$ the half side length of the corresponding square. The maximal value that $cx_i$ and $cy_i$ can respectively take is equal to half the side length $l$ of the corresponding square. In the same way, the minimal value, that $cx_i$ and $cy_i$ can respectively take, is equal to $-l$. The Pythagorean theorem is used to define $l$ from the radius $r$ of the circle. The latter verifies: 

\begin{center}
    \begin{small}
$l^2 + l^2 = r^2$\\
$l = \dfrac{r}{\sqrt{2}}$\\
\end{small}
\end{center}

Regarding the quadratic function, one solution consists to replace it by the Manhattan distance that uses the absolute norm. Today's automatic solvers are quite able to deal with absolute values. 
Therefore, the overall problem formulation can be rewritten as follow :

    \begin{align}
    &\min \qquad && \sum_{\forall (i,j) \in  E} x_{ij} \cdot \lvert cx_i - cx_j\rvert  + x_{ij} \cdot \lvert cy_i - cy_j\rvert \\ 
    & \text{s.t.} \qquad && \nonumber \\
    &&& cx_{i}\leq m_i + \dfrac{r_i}{\sqrt{2}} \qquad \qquad \forall i \in \mathcal{V} \label{MP11}\\
    &&& cx_{i}\geq m_i - \dfrac{r_i}{\sqrt{2}} \qquad \qquad \forall i \in \mathcal{V} \label{MP12}\\
    &&& cy_{i} \leq n_i + \dfrac{r_i}{\sqrt{2}} \qquad \qquad \forall i \in \mathcal{V} \label{MP13}\\
    &&& cy_{i} \geq n_i - \dfrac{r_i}{\sqrt{2}} \qquad \qquad \forall i \in \mathcal{V} \label{MP14}\\
    &&& (\ref{MP3}),(\ref{MP4}),(\ref{MP5}), (\ref{MP}) \\
    &&& x_{ij} \in \{0,1\}, u_{i} \in \mathbb{N}, (cx_i,cy_i) \in \mathbb{R}\times\mathbb{R} &&  \label{MP15}
\end{align}

The objective function minimizes the Manhattan distance between hitting points using arcs where $x_{ij}=1$, according to cycle found by the TSP constraints added at the end of the formulation. Then, constraints \ref{MP11}, \ref{MP12}, \ref{MP13}, and \ref{MP14} ensure that each of $cx_i$ and $cy_i$ belong to the interval $[-\dfrac{r}{\sqrt{2}}, \dfrac{r}{\sqrt{2}}]$ respectively. 

\subsection{CETSP mathematical formulation : Small hexagon}
\begin{figure}[ht]
\centering
     \includegraphics[scale=0.13]{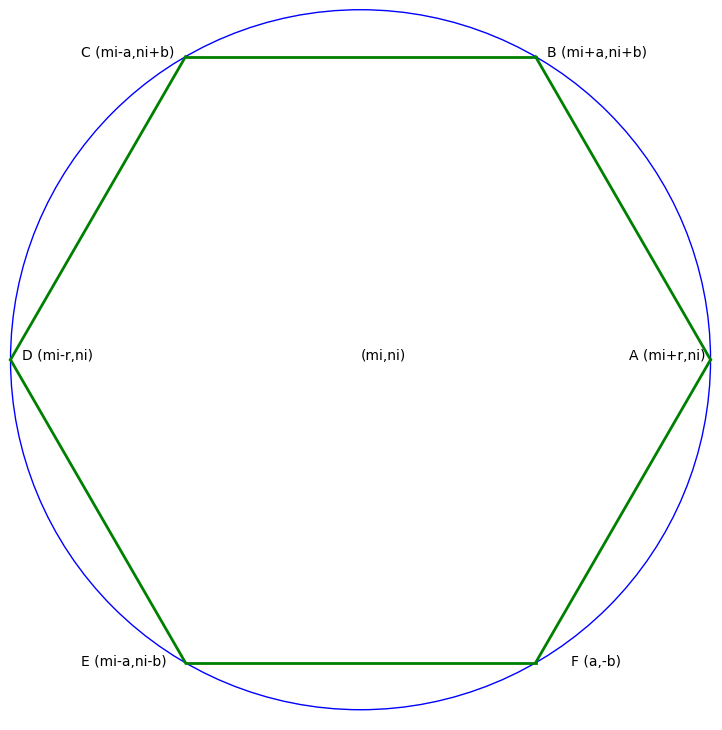}
     \caption{Inscribing a hexagon in a circle}
     \label{hexagoneincircle}
\end{figure}

The second model consists at inscribing a hexagon which is a closest form to that of the circle and allows to cover $76.98\%$ of its space. We propose a linear manner to do it by adding the six line equations as constraints to frame the search space of each sensor. Two points of each line are needed to define equations. Nevertheless, the number of points goes down to only six for each sensor, because each line shares two points with two other lines. Figure \ref{hexagoneincircle} shows the hexagon inside the circle and the coordinates for the six points ($A,B,C,D,E,F$). 

The coordinates of points $A(m_i+r,n_i)$ and $D(m_i-r,n_i)$ are easy to find using the radius. The coordinates of the other points $B(m_i+a,n_i+b)$, $C(m_i-a,n_i+b)$, $E(m_i-a,n_i-b)$, $F(m_i+a,n_i-b)$ are calculated from the values of $a$ and $b$ where :

\begin{center}
$a = r \cdot \cos(\frac{\pi}{3})$ \\
$b = r \cdot \sin(\frac{\pi}{3})$
\end{center}

Then, both the slope and the intercept are computed. The slope $s$ of each line is computed by considering two successive side points $(x_1,y_1)$ and $(x_2,y_2)$ 
as follows: 
\begin{center}
$ s = \frac{y_2 - y_1}{x_2 - x_1}$
\end{center}
Regarding the intercepts, one point $(x_1,y_1)$ and the slope $s$ are needed and used in this way:


\begin{center}
$intercepte = y_1 - s \cdot x_1$
\end{center}
This procedure is repeated for each side of the hexagon. Finally, six constraints are added to each sensor to define the search space. 

The complete CETSP mathematical formulation is defined as follows:


\begin{align}
    &\min \qquad && \sum_{\forall (i,j) \in  E} x_{ij} \cdot \lvert cx_i - cx_j\rvert  + x_{ij} \cdot \lvert cy_i - cy_j\rvert \label{MNH} \\ 
    & \text{s.t.} \qquad && \nonumber \\
    &&& slope_i^{AB} \cdot cx_{i} + intercepte_i^{AB} - cy_{i} \leq 0 \qquad \qquad \forall i \in \mathcal{V} \label{MP22}\\
    &&& slope_i^{CD} \cdot cx_{i} + intercepte_i^{CD} - cy_{i} \leq 0 \qquad \qquad \forall i \in \mathcal{V} \label{MP24}\\
    &&& slope_i^{DE} \cdot cx_{i} + intercepte_i^{DE} - cy_{i} \geq 0 \qquad \qquad \forall i \in \mathcal{V} \label{MP25}\\
    &&& slope_i^{FA} \cdot cx_{i} + intercepte_i^{FA} - cy_{i} \geq 0 \qquad \qquad \forall i \in \mathcal{V} \label{MP27}\\
    &&& cy_i \leq m_i + b  \qquad \qquad \qquad \qquad  \qquad \qquad \qquad \quad   \forall i \in \mathcal{V} \label{MP23}\\
    &&& cy_i \geq m_i - b \qquad \qquad \qquad \qquad  \qquad \qquad \qquad \quad  \forall i \in \mathcal{V} \label{MP26}\\
    &&& (\ref{MP3}),(\ref{MP4}),(\ref{MP5}), (\ref{MP}) \\
    &&& x_{ij} \in \{0,1\}, u_{i} \in \mathbb{N}, (cx_i,cy_i) \in \mathbb{R}\times\mathbb{R} &&  \label{MP28}
\end{align}

\section{Linearization of the objective function}
The objective function \ref{MNH} contains absolute values representing the differences between each two hitting points $i$ and $j$, on both the $x$ and $y$ axes. Two linearization options are proposed.

\subsection{Linearization 1}
Four temporary variables are used for each $(i,j)$ combination : $tx1, tx2, ty1, ty2$. The considered linearization is performed as follows:

\begin{center}
\begin{small}
$\min \quad \sum_{\forall (i,j) \in  E} x_{ij} \cdot \lvert cx_j - cx_i\rvert + x_{ij} \cdot \lvert cy_j - cy_i\rvert$\\
\end{small}
\end{center}
\begin{minipage}[t]{0.48\linewidth}
\begin{center} min $\lvert cx_j - cx_i\rvert$\end{center}
\begin{center}
\begin{small}
$\Downarrow$ \\
$\min \quad tx1_{ij} + tx2_{ij}$\\
$\qquad \qquad \qquad tx1_{ij} - tx2_{ij} = cx_j - cx_i$\\
$tx1_{ij} \geq 0$\\
$tx2_{ij} \geq 0$
\end{small}
\end{center}
\end{minipage}\hfill
\begin{minipage}[t]{0.48\linewidth}
\begin{center} min $\lvert cy_j - cy_i\rvert$\end{center}
\begin{center}
\begin{small}
$\Downarrow$ \\
$\min \quad ty1_{ij} + ty2_{ij}$\\
$\qquad \qquad \qquad ty1_{ij} - ty2_{ij} = cy_j - cy_i$\\
$ty1_{ij} \geq 0$\\
$ty2_{ij} \geq 0$
\end{small}
\end{center}
\end{minipage}

\vspace{4mm}
 
The absolute values by are linearized by adding the four positive numeric variables, previously mentioned, for each couple of sensors. Then, the used constraints ensure the positivity of each difference on the X or Y Cartesian axis. 

\subsection{Linearization 2}
Two changes are added to the previous model. First, the constant $M$ represents the largest possible distance (on X or Y axis) between two sensors. 
These latter variables force all possible configurations of the absolute value to be equal to the same positive value.

\begin{minipage}[t]{0.48\linewidth}
\begin{center} $min  \lvert cx_j - cx_i\rvert$\end{center}
\begin{center}
\begin{small}
$\Downarrow$ \\
$\min \quad d^x_{ij}$\\
$\qquad \qquad \qquad d^x_{ij} \geq cx_i - cx_j - M \cdot (1 - x_{ij})  $\\
$\qquad \qquad \qquad
d^x_{ij} \geq cx_j - cx_i - M \cdot (1 - x_{ij}) $
\end{small}
\end{center}
\end{minipage}\hfill
\begin{minipage}[t]{0.48\linewidth}
\begin{center} min $\lvert cy_j - cy_i\rvert$\end{center}
\begin{center}
\begin{small}
$\Downarrow$ \\
$\min \quad d^y_{ij}$\\
$\qquad \qquad \qquad d^y_{ij} \geq cy_i - cy_j - M \cdot (1 - x_{ij})  $ \\
$\qquad \qquad \qquad d^y_{ij} \geq cy_j - cy_i - M \cdot (1 - x_{ij})  $
\end{small}
\end{center}
\end{minipage}

The constraints  
indicate that $d^x_{ij}$ and $d^y_{ij}$ are equal to the X and Y distances between $i$ and $j$ respectively, in the case where $x_{ij }= 1$, otherwise it is equal to $0$. 

\section{Linear regression}
\begin{figure}[ht]
\centering
     \includegraphics[scale=0.30]{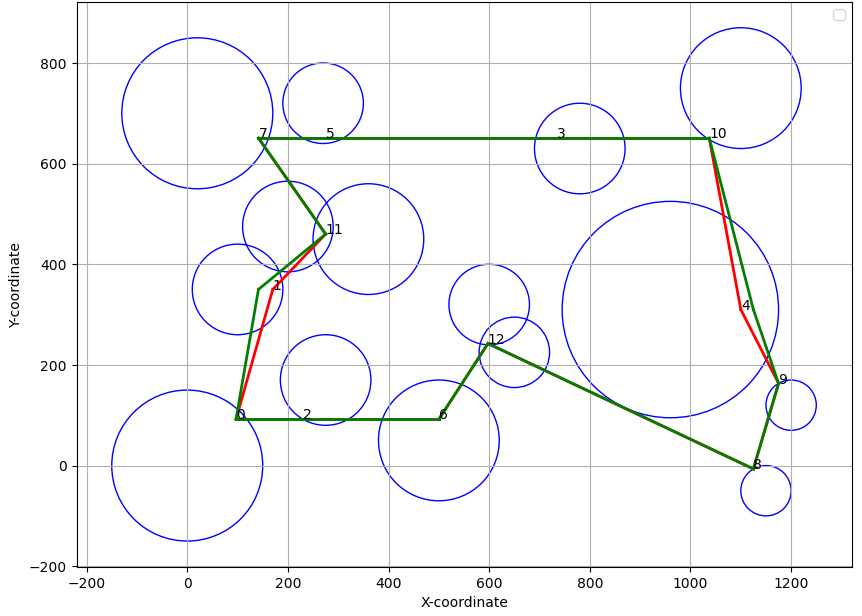}
     \caption{The Manhattan metric underestimates the Euclidean metric}
     \label{underestimateMHTD}
\end{figure}

After testing the Manhattan distance, we observed sometimes an underestimation, and other times an overestimation of the Euclidean distance. In order to find a mismatch between the two metrics, the CPLEX-based sparse approach is used to find an intermediate cycle as shown in yellow in Figure \ref{underestimateMHTD}. After trying to optimize the Manhattan distance for the two fragments $(0-1-11)$ and $(10-4-9)$, the trajectory deviated from green towards the red lines for two new hitting points. The hitting point numbered 1 was moved from the coordinates $(170,350)$ to $(141,547,350)$ for the same Manhattan distance $647,113$ but a worse Euclidean distance from $566.89$ to $573,405$. Similarly, the hitting point numbered 4 is moved from the coordinates $(1100,310)$ to $(1125,310)$ for the same Manhattan distance of $1047.42$ and, as before, with a worse Euclidean distance increasing from $816.997 $ to $834,568$.

Therefore, a reasonable choice would be to find an approximation of the Euclidean distance from the Manhattan distance and then check whether the latter brings an improvement to the travel cost.

The solution proposed is to learn from a list of, for example, $1000000$ absolute values for randomly generated Cartesian coordinates to predict their Euclidean distances. The goal of this learning is to find coefficients for the two absolute input values allowing the Euclidean distance to be approached as closely as possible.

\begin{figure}[ht]
\centering
     \includegraphics[scale=0.6]{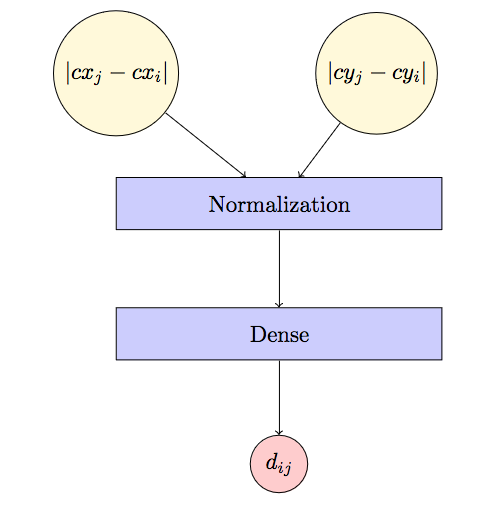}
     \caption{The architecture of the neural network used}
     \label{NNArchitecture}
\end{figure}

A very simple neural network was used as illustrated in Figure \ref{NNArchitecture}. The latter consists of a layer for normalizing the input parameters and a second dense layer receiving the normalized inputs and generating a single output.

\begin{figure}[ht]
\centering
     \includegraphics[scale=0.30]{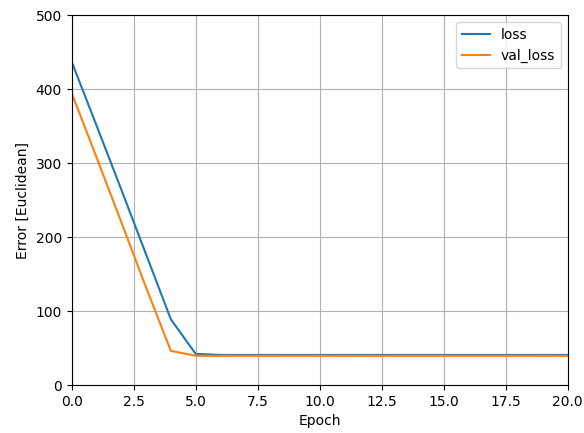}
     \caption{The loss function evolution}
     \label{Losscurve}
\end{figure}

Figure \ref{Losscurve} shows the evolution of the loss function (error) during the epochs (iterations) of the learning process in blue and the evaluation one in orange. The loss represents the difference between the approximation found and the actual Euclidean distance. The error decreases well during the learning and evaluation iterations.

The coefficients obtained after learning have the following values:
\begin{itemize}
    \item The coefficient for the difference on the x axis in absolute value:  
    \begin{center}
        $C_{dx} = (170.98 / 235.656533)$
    \end{center} 
    \item The coefficient for the difference on the y axis in absolute value :
    \begin{center}
    $C_{dy} = (168.928 / 235.695644)$
    \end{center}
    \item The bias: 
    \begin{center}
    $Bias = 503.279$
    \end{center}
\end{itemize}

The new objective function will have the following form: 

 $\min \qquad  \sum_{\forall (i,j) \in  E} x_{ij} \cdot (C_{dx} \cdot \lvert cx_i - cx_j\rvert - C_{dx} \cdot  mean_{dx} + C_{dy} \cdot \lvert cy_i - cy_j\rvert - C_{dy} \cdot mean_{dy} + Bias)  $

\section{Vector projections}
The technique employed and discussed in this section is inspired by \cite{camino2019linearization}, as outlined in their work.
In an aim of enhancing the optimization and approximating further the Euclidean distance, the computed distance between each couple of sensors is projected on eight new axes. These axis divide the Cartesian space into eight equal parts and are spaced by an angle of $\pi$. The first added axe makes an angle of $\pi$ with the vector relating the two considered sensors. Then, the second axis makes an angle of $2\pi$ with the distance vector and so on for the rest of the added axis.

\begin{figure}[ht]
\centering
     \includegraphics[scale=0.4]{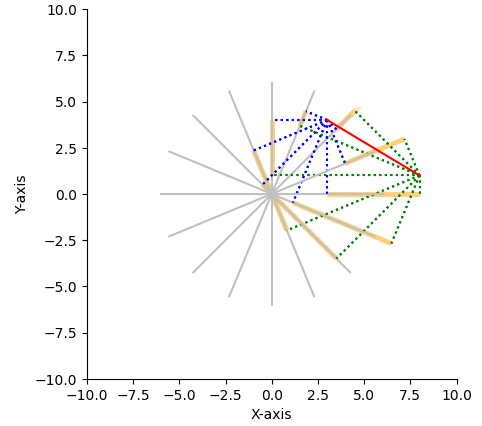}
     \caption{The projection of the distance on eight new axis}
     \label{projection}
\end{figure}

These projections help us find out the length of the distance vector on eight more axis rather than only two (X an Y axis) as shown in Figure \ref{projection}. After calculating the projections (orange parts of each axis), their sum is added to the original objective function and is minimized.

This approach essentially utilizes the trigonometric properties of the cosine and sine functions to determine the direction of the axis relative to the red line and then calculates the projection accordingly. It begins by defining the Eight Axes originating from the origin. Each axis direction is defined by a unit vector obtained from the cosine and sine of the angles $i\cdot8\pi$, where $i$ ranges from 1 to 8. Then, the projection is computed using the dot product. Mathematically, the steps for projecting each of the two points forming the red line can be represented as follows:

\begin{enumerate}
\item Calculate the Unit Vector representing the direction of axis $i$ using the formula:
\begin{center}  
$unit\_axis=(cos(i\cdot8\pi),sin(i\cdot8\pi))$.
\end{center}
\item Compute the dot product between the original point $P$ and the unit axis:
\begin{center} 
$\text{dot\_product} = \vec{P} \cdot \text{unit\_axis}$ .
\end{center}
\item Calculate the projection $P'$ using the dot product and the unit vector representing axis $i$:
\begin{center} 
$P' = \text{dot\_product} \times \text{unit\_axis}$.
\end{center}
\end{enumerate}

\section{Fragmented solver-based approach (MF)}
The approach starts by detecting the groups of intersecting disks over the sensors. Then, a submodel is applied to each group $\mathcal{z}$ in order to find out the coordinates $(cx,cy)$ of the point present in the intersection zone. This point should reduce the average distance over all the sensors that are concerned by the intersection, each of coordinate $(x_k,y_k)$. Therefore, this submodel called $P^{Intersection}_{CETSP}$ is applied to each $k$ and is  described as follows :

\begin{align}
     & \min  \qquad && \sum_{\forall k \in \mathcal{Z}} \lvert cx - x_{k}\rvert +\lvert cy - y_{k}\rvert  \\
    & \text{s.t.} \qquad && \nonumber \\
        &&& (\ref{MP3}),(\ref{MP4}),(\ref{MP5}), (\ref{MP}) && \\
        &&& (9)-(12)/(16)-(21) \label{CV2}\\
&&& x_{ij} \in \{0,1\} \forall (i,j) \in E, \quad u_i \in \mathbb{N} \forall i \in V, (cx_i,cy_i) \in \mathbb{R}\times\mathbb{R} \forall i \in V   \label{MP9}
\end{align}

Given the distance matrix $d_{ij}$ between each two centre points $i$ and $j$, an automatic solver is used the Hamiltonian cycle connecting the sensors exactly as a TSP problem. Then, a loop through the sensors $i$ respecting the found Cycle order is carried out in order to optimize the travel cost over the actual chunk $\{i-1,i,i+1\}$ using this small model $P^{GEO}_{CETSP}$ :
\begin{align}
     & \min \qquad && \lvert cx_i - x_{i-1}\rvert +\lvert cy_i - y_{i-1}\rvert + \lvert cx_i - x_{i+1}\rvert +\lvert cy_i - y_{i+1}\rvert \\
    & \text{s.t.} \qquad && \nonumber \\  
        &&& (\ref{MP3}),(\ref{MP4}),(\ref{MP5}), (\ref{MP}) && \\
        &&& (9)-(12)/(16)-(21) \label{CV}\\
&&& x_{ij} \in \{0,1\} \forall (i,j) \in E, \quad u_i \in \mathbb{N} \forall i \in V, (cx_i,cy_i) \in \mathbb{R}\times\mathbb{R} \forall i \in V   \label{MP9_1}
\end{align}

In the same loop, an other model is used to optimize the travel cost for the actual chunk $\{i-2,i,i+2\}$. The latter is described as follows $P^{GEO1}_{CETSP}$ :

\begin{align}
     & \min \qquad && \lvert cx_i - x_{i-2}\rvert +\lvert cy_i - y_{i-2}\rvert + \lvert cx_i - x_{i+2}\rvert +\lvert cy_i - y_{i+2}\rvert \\
    & \text{s.t.} \qquad && \nonumber \\
        &&& (\ref{MP3}),(\ref{MP4}),(\ref{MP5}), (\ref{MP}) && \\
        &&& (9)-(12)/(16)-(21) \label{CV1}\\
&&& x_{ij} \in \{0,1\} \forall (i,j) \in E, \quad u_i \in \mathbb{N} \forall i \in V, (cx_i,cy_i) \in \mathbb{R}\times\mathbb{R} \forall i \in V   \label{MP9_2}
\end{align}

These kind of sub-models are easier to solve by the model as the number of variables and constraints decreases, which is better, especially, for big problems where the number of sensors increases. After solving each sub-model, a matrix for hitting points is updated by the new coordinates. This matrix was initialized by the center points coordinates. 

Basically, this fragmented CPLEX-based approach is composed of three major components; managing intersecting zones using $P^{Intersection}_{CETSP}$ to define the list of super-nodes. Then, the TSP problem is solved considering the super-nodes to find out the best order of visit. Finally, each hitting point is relocated inside the respective emission areas to reduce the travel cost further using $P^{GEO}_{CETSP}$ and $P^{GEO1}_{CETSP}$.

The steps through these components, carried out on each fragment (chunk of the route), are described in Algorithm \ref{algoMF}. 

\begin{algorithm}[!htb]
{\bf Input.} $N$, $(cx_i,cy_i) \forall i \in \mathcal{S}$ \\
{\bf Output.} $(hx_i,hy_i) \forall i \in \mathcal{S}$, $route[N]$.
\caption{MF algorithm}\footnotesize
\label{algoMF}
\begin{algorithmic}[1]
  \STATE {Set $C_{dx}, C_{dy}, B$.}
  \STATE {Read the convex set $C^i$.}
  \STATE {Read the linearization option $L$.}
  \STATE {$d_{ij} = \sqrt{(cx_i - cx_j)^2 + (cy_i - cy_j)^2} \quad \forall (i,j) \in \mathcal{E}$}
  \STATE {Detect the groups of intersections $\mathcal{H}$;}
  
  \FORALL{ $Z \in \mathcal{H}$}
  \STATE {$(cx_i,cy_i)^Z$ = $P^{Intersection}_{CETSP}$ }
  \ENDFOR
  \STATE {$Ncost,route[N] = TSP();$}
  \STATE {$Ocost = 1200000; b = 1;$}
  \WHILE{$b == 1$}
  \STATE {$b = 0$}
  \WHILE{$Ocost \geq Ncost$}
  \FORALL{ $\omega$ of $\mathcal{S}$}{
  \STATE {$Ncost,(hx_\omega,hy_\omega) = P^{GEO}_{CETSP}$}
  \STATE {$Ncost,(hx_\omega,hy_\omega) = Ubiquit()$}
  }
  \ENDFOR
  \FORALL{ $\omega$ of $\mathcal{S}$}
  \STATE {$Ncost,(hx_\omega,hy_\omega) = P^{GEO1}_{CETSP}$}
  \STATE {$Ncost,(hx_\omega,hy_\omega) = Ubiquit()$}
  \ENDFOR
  \ENDWHILE
  \STATE {$cost,route[N]  = TSP()$}
  \IF{$cost \geq Ncost$} 
  \STATE {$Ncost = cost$}
  \STATE {$b = 1$}
  \ENDIF
  \ENDWHILE
  \STATE {$Ncost,route[N] = TSP();$}
\end {algorithmic}
\end {algorithm}

It needs as an input the number of sensors $N$ and the Cartesian coordinates $(cx_i,cyi)$ of each one of them $i$. The algorithm is supposed to compute the solution to the problem by giving the hitting points coordinates $(hx_i,hyi)$ and the order of visit of each hitting point presented in a vector $route[N]$ of size $N$.

The algorithm starts by initializing the linear regression coefficients used for approximating the Euclidean distance and choosing which convex set $C^i$ to consider and which linearization option $L$ to use ($Lin1$ or $Lin2$). Then, the Euclidean distance matrix is calculated for each couple of sensors using the inputted coordinates. This latter is going to be used in all components of the code. After preparing all parameters, the first component of the approach begins by determining the groups of intersecting emission areas $\mathcal{H}$ to be able to define one super-node for each group $\mathcal{Z} \in \mathcal{H}$ using $P^{Intersection}_{CETSP}$. Now that the list of super-nodes is ready, the routing $route[N]$ with the optimal travel cost $Ncost$ is found using $TSP()$. The statement 10 of the code set a fairly large number to the variable $Ocost$ that will represent the old cost before optimization in the next loop and a boolean variable is set to true ($1$) to be able to enter the first while loop. The second while loop is carried out while the travel cost is still improving due to the relocation statements inside of it. $P^{GEO}_{CETSP}$ is applied for each super-node $\omega$, to reduce the cost of the routing going from $\omega - 1$ to $\omega + 1$, followed by $Ubiquit()$ to reduce the distance between the actual hitting point and the intersection point between two lines the first relating the precedent and next sensor with respect to the $route[N]$ vector, and the line relating the hitting point and perpendicular on the first one. Then, $P^{GEO1}_{CETSP}$ is carried out at the same way. These final steps produce updated hitting points, which are subsequently utilized by the TSP model to validate the order of visits for the new points.

\section{Computational Results}
Instances of various sizes, ranging from small to large, have been randomly selected from a real-world use case. Specifically, there are five small instances consisting of 10 sensors each, ten medium instances consisting of 15 or 20 sensors (ns15nb0 to ns20nb4), and five large instances consisting of 50 sensors (ns50nb0 to ns50nb4). These instances are intended to assess the effectiveness of the proposed mathematical formulations for different convex sets and the two proposed linearizations.

\subsection{The assessment of the different convex sets}
The diverse proposed shapes are evaluated across all utilized instances. Table \ref{EucShapesTableAllInstEuc} presents the incurred cost in Euclidean metric and the corresponding computation time for solutions directly obtained through the automated solver CPLEX, while minimizing the Manhattan distance (MD).

\begin{table}[htb] \tiny
\caption{\scriptsize{Comparison between the two convex sets (The numbers in red correspond to the shortest distance with regard to the Euclidean metric.).}}
\begin{center}
    \begin{tabular}{|c|c|c|c|c|c|c|c|c|}
    \hline
    Instances & \multicolumn{3}{c|}{Small square} & \multicolumn{3}{c|}{Small hexagon}  \\
   & MD & Cost (m) & Time (s) & MD & Cost (m)  & Time (s)  \\
  \hline
  ns10nb0 & 3203.36 & 2549.47 &  0  &  3147.64 & \textcolor{red}{2520.33} &  0   \\
  ns10nb1 & 2954.71 &  \textcolor{red}{2545.52} &  0 & 2896.99 & 2637.74  &  0   \\
  ns10nb2 & 3858.91 & 3476.29 &  0 & 3772.71 &  \textcolor{red}{3408.68} &   0   \\
  ns10nb3 & 4212.34 & 3774.33 &  1  & 4099.45 &  \textcolor{red}{3428.36} &  2   \\
  ns10nb4 & 3030.57 & \textcolor{red}{2520.14}  &  0  & 2969.85 & 2532.57  &  1   \\
  ns15nb0 & 3007.9 & 2526.51 &  16  &  2831.03 & \textcolor{red}{2466.91}  &  44   \\
  ns15nb1 & 3044.91 & 2467.18 &  5  & 2928.65 &  \textcolor{red}{2323.45} &  20    \\
  ns15nb2 & 3196.59 & 2735.53 &  10  & 3103.38 &  \textcolor{red}{2700.61}  &  39   \\
  ns15nb3 & 5140.83 & 4516.01 &  4  & 5085.06 &  \textcolor{red}{4320.94}  &  16   \\
  ns15nb4 & 4411.27 & \textcolor{red}{3676.96} &   6  & 4338.86  & 3694.71 &   30   \\
  ns20nb0 & 3405.19 & 2916.96  &  313  & 3164.05 &  \textcolor{red}{2864.38} &   320   \\
  ns20nb1 & 4183.68 & \textcolor{red}{3377.03}  &  321  & 4007.83 &  3436.18 &   328   \\
  ns20nb2 & 4507.74 & 4069.68 &  34  & 4244.01 &  \textcolor{red}{3741.06} &  129   \\
  ns20nb3 & 5246.94 & 4786.6 &  107  & 5050.3 & \textcolor{red}{4630.23} &   241   \\
  ns20nb4 & 4369.21 & 3912.39  &  310  &  4211.33 &  \textcolor{red}{3905.93} &   325   \\
  ns50nb0 & 10509.5 & 8862.15  &  403  & 10064.6 &  \textcolor{red}{8617.52} &   334  \\ 
  ns50nb1 & 10131.5 & 8134.57  &  387  & 8998.93 &  \textcolor{red}{7796.04} &   366  \\ 
  ns50nb2 & 9755.11 & \textcolor{red}{8141.69} &   300  & 10403.8 & 8736.35 &   256   \\ 
  ns50nb3 & 8548.1 & \textcolor{red}{7276.82} &  335  & 8619.19 & 7357.58 &  341   \\
  ns50nb4 & 9479.88 &  \textcolor{red}{7778.93} &  357  & 9962.29 &  8152.05 & 328   \\
  ns70nb0 & 15840.2 & \textcolor{red}{12646.2} & 280 & 31124.3 & 24997.2 &  255   \\
  \hline
  N°best costs \& Average Time & / & 8 & 151.8571 & / & 13 & 160.7143  \\ 
  \hline
\end{tabular}
\end{center}
\label{EucShapesTableAllInstEuc}
\end{table}

Observing the table, we can conclude that the small hexagon yields a better travel cost 
for almost all small and medium sized instances, with reasonable computation time. The latter is relatively similar for small instances, but gradually slow down for medium instances. On the other hand, the small square formulation outperforms the small hexagon for larger instances. This discrepancy can be explained by the fact that the search space for hexagons becomes enormous, which interrupts the solver before it manages to exhaustively explore the entire search space and stops, which is why the difference in reported calculation times does not seem significant, compared to the small square formulation.

The final row of the table indicates the average calculation time required by the solver to achieve the specified travel cost. Remarkably, this time appears to be reasonably proportional to the size of the search space.

\subsection{Comparison of linearizations}
The two models linearizing the objective function are tested on the instances considered for all the convex sets defined in this article: the small square and the small hexagon. The results of this comparison are illustrated in Tables \ref{LinAbsValObjSSquareCplex} and \ref{LinAbsValObjSHexCPLEX} when the overall model is solved by CPLEX on small squares and small hexagons respectively. However, Tables \ref{LinAbsValObjSSquareMF} and \ref{LinAbsValObjSHexMF} show the results found by the fragmented method (MF) when small squares and small hexagons are considered respectively.

For each of the tables below, the first column presents the names of the instances, and the second column (VA) presents the Euclidean cost found before any linearization. Then, the third and fourth columns are multicolumns, of size three, reporting the first and second types of linearizations respectively. Each multicolumn displays the Manhattan and Euclidean distances, as well as the CPU calculation time took to find each found solution. The best Euclidean travel cost between linearizations is shown in red. When the Euclidean cost found before linearization 1 is lower than the one found by the absolute value formulation, a small star is added.

The displacement cost, given by linearization 2, when using CPLEX, is worse than that given when the calculation time was limited to around 20 seconds. Therefore, the best solutions are those reported in the tables.

\begin{table}[htb] \tiny
\caption{\scriptsize{Comparison of the two linearizations considering small squares (CPLEX)}}
\begin{center}
    \begin{tabular}{|c|c|c|c|c|c|c|c|c|}
    \hline
  & VA & \multicolumn{3}{c|}{Linearization 1} & \multicolumn{3}{c|}{Linearization 2}  \\
  Instances & & Manhattan  & Euclidean & T & Manhattan & Euclidean & T \\
  \hline
ns10nb0 & 2549.47* & 3328 & \textcolor{red}{2654} & 7  & 3836.13 & 3086.33 &   12  \\ 
ns10nb1 & 2545.52 & 2954.71 & 2509.1 & 80 & 2954.71 &  \textcolor{red}{2482.45} & 21  \\ 
ns10nb2 & 3476.29* & 4199.3 & \textcolor{red}{3558.78} & 57 & 5021.17 & 3979.06 &   7  \\ 
ns10nb3 & 3774.33 & 4230.97 & \textcolor{red}{3695.41} & 57 & 4212.34 &  3700.55 &   20  \\ 
ns10nb4 & 2520.14 & 3030.57 & 2478.25 & 82  & 3030.57 &  \textcolor{red}{2347.23} &  20  \\ 
ns15nb0 & 2526.51* & 3007.9 & \textcolor{red}{2546.11} & 78  & 3007.9 & 2548.96 & 16  \\ 
ns15nb1 & 2467.18 & 3044.91 & 2439.23 & 83  & 3044.91 & \textcolor{red}{2428.83} & 16  \\ 
ns15nb2 & 2735.53 & 3196. & \textcolor{red}{2709.85} & 75  & 3218.15 &  2725.63 & 21  \\ 
ns15nb3 & 4516.01 & 5140.83 & \textcolor{red}{4340.38} & 77  & 5147.73 & 4343.03 & 22 \\ 
ns15nb4 & 3676.96* & 4411.27 & 3865.65 & 76  & 4411.27 &  \textcolor{red}{3825.95} & 18  \\ 
ns20nb0 & 2916.96 & 3533.95 & 2841.44 & 58  & 3536.67 & \textcolor{red}{2739.71} & 17  \\ 
ns20nb1 & 3377.03 & 4183.68 & \textcolor{red}{3298.78} & 56 & 4183.69 & 3308.84 & 16  \\ 
ns20nb2 & 4069.68 & 4507.74 & 3821.17 & 68  & 4529.3 & \textcolor{red}{3698} &18 \\ 
ns20nb3 & 4786.6 & 5246.94 & 4658.17 & 72 & 5246.94 &  \textcolor{red}{4609.92} & 18 \\ 
ns20nb4 & 3912.39* & 4335.23 & \textcolor{red}{3921.56} & 62  & 4554.99 & 3931.43 & 16 \\ 
\hline
\end{tabular}
\end{center}
\label{LinAbsValObjSSquareCplex} 
\end{table}

\begin{table}[!htb] \tiny
\caption{\scriptsize{Comparison of the two linearizations considering small squares (MF)}}
\begin{center}
    \begin{tabular}{|c|c|c|c|c|c|c|c|c|c|c|c|}
    \hline
  & \multicolumn{3}{c|}{VA} & \multicolumn{3}{c|}{Linearization 1} & \multicolumn{3}{c|}{Linearization 2}  \\
  Instances & Manhattan  & Euclidean (m) & T (s) & Manhattan  & Euclidean (m) & T (s) & Manhattan & Euclidean (m) & T (s)\\
  \hline
ns10nb0.txt & 3251.82 & 2569.22 & 0 &  3305.02 & 2592.52 & 0 &  3200.05 & \textcolor{red}{2552.47} & 0 \\ 
ns10nb1.txt & 3106.73 & \textcolor{red}{2409.21} & 0 &  3179.87 & 2472.31 & 0 &  3090.16 & 2413.92 & 0 \\ 
ns10nb2.txt & 3978.83 & \textcolor{red}{3126.72} & 0 &  4104.78 & 3215.57 & 0 &  3978.83 & 3146.29 & 0 \\ 
ns10nb3.txt & 4282.78 & \textcolor{red}{3340.37} & 0 &  4389.92 & 3414.66 & 0 &  4282.78 & \textcolor{red}{3340.61} & 0 \\ 
ns10nb4.txt & 3159.6 & 2390.68 & 0 &  3237.14 & 2410.03 & 0 &  3129.99 & \textcolor{red}{2330.85} & 0 \\ 
ns15nb0.txt & 3120.76 & 2531.48 & 0 &  3208.47 & 2475.91* & 0 &  3120.76 & \textcolor{red}{2441} & 0 \\ 
ns15nb1.txt & 3066.86 & 2357.18 & 0 &  3066.86 & 2356.35* & 0 &  3032.04 & \textcolor{red}{2351.23} & 0 \\ 
ns15nb2.txt & 3222.06 & 2638.5 & 0 &  3350.06 & 2689.22 & 0 &  3205.49 & \textcolor{red}{2607.84} & 0 \\ 
ns15nb3.txt & 5661.2 & \textcolor{red}{4275.8} & 0 &  5750.9 & 4325.25 & 0 &  5661.2 & 4286.27 & 0 \\ 
ns15nb4.txt & 4639.86 & 3546.99 & 0 &  4686.6 & 3580.06 & 0 &  4557.01 & \textcolor{red}{3508.48} & 0 \\ 
ns20nb0.txt & 3561.28 & \textcolor{red}{2660.4} & 0 &  3642.13 & 2732.33 & 0 &  3568.99 & 2673.4 & 0 \\ 
ns20nb1.txt & 4148.47 & \textcolor{red}{3200.36} & 0 &  4413.56 & 3399.1 & 0 &  4169.3 & 3256.11 & 0 \\ 
ns20nb2.txt & 4612.32 & 3773.71 & 2 &  4740.18 & 3785.33 & 2 &  4636.26 & \textcolor{red}{3725.22} & 2 \\ 
ns20nb3.txt & 5460.73 & 4301.64 & 0 &  5814.84 & 4530.38 & 0 &  5426.52 & \textcolor{red}{4257.57} & 0 \\ 
ns20nb4.txt & 4513.85 & 3777.8 & 2 &  4678.82 & 3836.63 & 0 &  4359.87 & \textcolor{red}{3671.67} & 0 \\ 
ns50nb0.txt & 6257.01 & 5202.03 & 4 &  6629.68 & 5308.99 & 3 &  6255.9 & \textcolor{red}{5066.96} & 2 \\ 
ns50nb1.txt & 6639.63 & 5272.45 & 18 &  6629.21 & 5171.64* & 27 &  6589.43 & \textcolor{red}{5159.29} & 19 \\ 
ns50nb2.txt & 5866.2 & 4777.01 & 5 &  6151.32 & 4897.7 & 2 &  5897.93 & \textcolor{red}{4764.3} & 1 \\ 
ns50nb3.txt & 6080.24 & 4976.47 & 10 &  6335.07 & 5119.89 & 2 &  6021.42 & \textcolor{red}{4797.71} & 2 \\ 
ns50nb4.txt & 7275.41 & 5526.13 & 10 &  7475.21 & 5659.45 & 8 &  7258.89 & \textcolor{red}{5524.44} & 7 \\ 
ns70nb0.txt & 7979.58 & 6379.79 & 134 &  7941.93 & 6238.18* & 309 &  7793.16 & \textcolor{red}{6155.21} & 1059 \\ 
  \hline
\end{tabular}
\end{center}
\label{LinAbsValObjSSquareMF} 
\end{table}

From Table \ref{LinAbsValObjSSquareMF}, it is noticed that Linearization 1 is worse than the absolute formulation and Linearization 2 at the same time, because it didn't provide the best performance on any of the tested instances even if it was able to improve the absolute value formulation of 4 instances, which represents $19.04\%$ of them, of about $74.705$ on average. Furthermore, despite the fact that the absolute value formulation is competitive to better than Linearization 2 for small instances, Linearization 2 gives the best solutions for almost all medium instances and the overall big instances. Basically, Linearization 2 improved $71.428\%$ of the tested instances of about $73.789$ on average.

\begin{table}[!htb] \tiny
\caption{\scriptsize{Comparison of the two linearizations considering small hexagons (CPLEX)}}
\begin{center}
    \begin{tabular}{|c|c|c|c|c|c|c|c|c|}
    \hline
  & VA & \multicolumn{3}{c|}{Linearization 1} & \multicolumn{3}{c|}{Linearization 2}  \\
  Instances & & Manhattan  & Euclidean (m) & T (s) & Manhattan & Euclidean (m) & T (s)\\
  \hline
ns10nb0 & 2520.33* & 3328 & 2654 & 7  & 3147.64 &  \textcolor{red}{2640.67} &  19 \\ 
ns10nb1 & 2637.74 & 2954.71 & 2509.1 & 80 &  2896.99 &  \textcolor{red}{2449.77} &  24 \\ 
ns10nb2 & 3408.68* & 4199.3 & \textcolor{red}{3558.78} & 57 & 5223.6 &  4392.08 &  1 \\ 
ns10nb3 & 3428.36* & 4230.97 & 3695.41 & 57 & 4119.85 &  \textcolor{red}{3457.82} &  20 \\ 
ns10nb4 & 2532.57 & 3030.57 & 2478.25 & 82  & 2969.85 &  \textcolor{red}{2275.78} &  24 \\ 
ns15nb0 & 2466.91* & 3007.9 & 2546.11 & 78  & 2831.03 &  \textcolor{red}{2468.08} &  19 \\ 
ns15nb1 & 2323.45* & 3044.91 & 2439.23 & 83  & 2928.65 &  \textcolor{red}{2348.02} &  19 \\ 
ns15nb2 & 2700.61* & 3196. & \textcolor{red}{2709.85} & 75  & \textcolor{red}{3103.38} &  2729.25 &  21 \\ 
ns15nb3 & 4320.94 & 5140.83 & 4340.38 & 77  & 5085.06 &  \textcolor{red}{4318.33} &  23 \\ 
ns15nb4 & 3694.71* & 4411.27 & 3865.65 & 76  & 4361.74 &  \textcolor{red}{3708.76} &  20 \\ 
ns20nb0 & 2864.38 & 3533.95 & 2841.44 & 58  & 3272.17 &  \textcolor{red}{2701.82} &  18 \\ 
ns20nb1 & 3436.18 & 4183.68 & 3298.78 & 56 & 4007.83 &  \textcolor{red}{3157.06} &  19 \\ 
ns20nb2 & 3741.06 & 4507.74 & 3821.17 & 68  & 4244.01 &  \textcolor{red}{3634.18} &  18 \\ 
ns20nb3 & 4630.23 & 5246.94 & 4658.17 & 72 & 5050.3 &  \textcolor{red}{4479.34} &  19 \\ 
ns20nb4 & 3905.93 & 4335.23 & 3921.56 & 62  & 4260.63 &  \textcolor{red}{3733.97} &  23 \\  
\hline
\end{tabular}
\end{center}
\label{LinAbsValObjSHexCPLEX} 
\end{table}

\begin{table}[!htb] \tiny
\caption{\scriptsize{Comparison of the two linearizations considering small hexagons (MF)}}
\begin{center}
    \begin{tabular}{|c|c|c|c|c|c|c|c|c|c|c|c|}
    \hline
  & \multicolumn{3}{c|}{VA} & \multicolumn{3}{c|}{Linearization 1} & \multicolumn{3}{c|}{Linearization 2}  \\
  Instances & Manhattan  & Euclidean (m) & T (s) & Manhattan  & Euclidean (m) & T (s) & Manhattan & Euclidean (m) & T (s)\\
  \hline
ns10nb0.txt & 3249.42 & \textcolor{red}{2561.53} & 0 &  3308.26 & 2587.58 & 0 &  3242.96 & 2568.88 & 0 \\ 
ns10nb1.txt & 2987.01 & 2402.83 & 0 &  3200.29 & 2485.38 & 0 &  2947.01 & \textcolor{red}{2370.38} & 0 \\ 
ns10nb2.txt & 3939.67 & 3134.86 & 0 &  4027.71 & 3208.47 & 0 &  3947.71 & \textcolor{red}{3119.68} & 0 \\ 
ns10nb3.txt & 4244.59 & 3302.16 & 0 &  4406.48 & 3422.48 & 0 &  4224.05 & \textcolor{red}{3291.44} & 0 \\ 
ns10nb4.txt & 3058.91 & 2329.69 & 1 &  3152.12 & 2362.49 & 0 &  3057.7 & \textcolor{red}{2290.83} & 0 \\ 
ns15nb0.txt & 2994.75 & \textcolor{red}{2423.85} & 0 &  3194.75 & 2480.63 & 0 &  3034.75 & 2439.94 & 0 \\ 
ns15nb1.txt & 2970.47 & 2328.57 & 0 &  3083.43 & 2340.51 & 0 &  2970.47 & \textcolor{red}{2300.3} & 0 \\ 
ns15nb2.txt & 3163.33 & 2590.93 & 1 &  3426.77 & 2725.92 & 0 &  3125.26 & \textcolor{red}{2553.88} & 0 \\ 
ns15nb3.txt & 5565.47 & \textcolor{red}{4206.61} & 0 &  5754.76 & 4324.81 & 0 &  5565.47 & 4210.51 & 0 \\ 
ns15nb4.txt & 4468.42 & \textcolor{red}{3495.34} & 0 &  4730.18 & 3597.25 & 0 &  4428.42 & 3499.17 & 0 \\ 
ns20nb0.txt & 3411.64 & 2589 & 1 &  3731.94 & 2807.72 & 0 &  3411.64 & \textcolor{red}{2587.97} & 0 \\ 
ns20nb1.txt & 4034.4 & \textcolor{red}{3168.69} & 1 &  4382.92 & 3367.37 & 0 &  4087.68 & 3183.05 & 0 \\ 
ns20nb2.txt & 4375.67 & \textcolor{red}{3550.52} & 2 &  4891.9 & 3827.54 & 3 &  4389.47 & 3611.85 & 2 \\ 
ns20nb3.txt & 5360.65 & 4232.54 & 1 &  5751.95 & 4474.3 & 0 &  5255.3 & \textcolor{red}{4216.64} & 0 \\ 
ns20nb4.txt & 4397.22 & 3723.18 & 2 &  4629.34 & 3778.36 & 0 &  4318.33 & \textcolor{red}{3652.86} & 0 \\ 
ns50nb0.txt & 6122.39 & \textcolor{red}{5016.17} & 9 &  6642.92 & 5241.18 & 5 &  6200.96 & 5142.52 & 3 \\ 
ns50nb1.txt & 6437.96 & 5165.15 & 23 &  7106.72 & 5546.78 & 23 &  6336.82 & \textcolor{red}{5038.75} & 23 \\ 
ns50nb2.txt & 5830.55 & 4684.69 & 8 &  6262.18 & 4868.68 & 4 &  5717.04 & \textcolor{red}{4655.77} & 2 \\ 
ns50nb3.txt & 6036.14 & 4898.62 & 4 &  6167.24 & 4813.3* & 7 &  5924.36 & \textcolor{red}{4764.53} & 1 \\ 
ns50nb4.txt & 7107.39 & \textcolor{red}{5431.4} & 9 &  7557.95 & 5675.87 & 6 &  7123.6 & 5432.24 & 6 \\ 
ns70nb0.txt & 7805.62 & 6204.97 & 144 &  8022.94 & 6229.37 & 156 &  7737.19 & \textcolor{red}{6116.44} & 228 \\ 
\hline
\end{tabular}
\end{center}
\label{LinAbsValObjSHexMF} 
\end{table}

From Table \ref{LinAbsValObjSHexMF}, it is noticed that Linearization 1 still does'nt give the best solutions for small hexagone as well. Moreover, Linearization 1 improves only one instance among the 21 tested instances. However, Linearization 2 enhances the travel cost for $61.904\%$ of instances of about $48.286$ on average and fails to enhance $38.095\%$ of the instances. 

In terms of computational time, there is no difference noticed for almost all the instances either when considering small squares or small hexagons, where it is a tiny amount of time compared to the CPLEX version, except for the bigger use case of $70$ sensors.

From these tables, it is noticed that the costs found by the two linearizations are competitive when using CPLEX on small squares, 
although the instances with a very large number of sensors could not be solved due to the limited capabilities of CPLEX towards the increase in the number of variables. However, the costs are more in favor of linearization 2 when using the fragmented CPLEX-based method. Furthermore, linearization 2 clearly gives better costs, for most instances whether using CPLEX or the fragmented method on small hexagons. Furthermore, the two linearizations did not slow down the CPLEX solution process, so the time remains reasonable. On the other hand, the fragmented method always takes less calculation time.

Despite the fact that Linearization 1 was able to guide the drone also inside the emission fields discs, which wasn't the case before linearization, it fails to improve the displacement cost of the majority of instances. Hence, this latter formulation will be discarded for the next experiments.

\begin{figure}[ht]
\centering
     \includegraphics[scale=0.25]{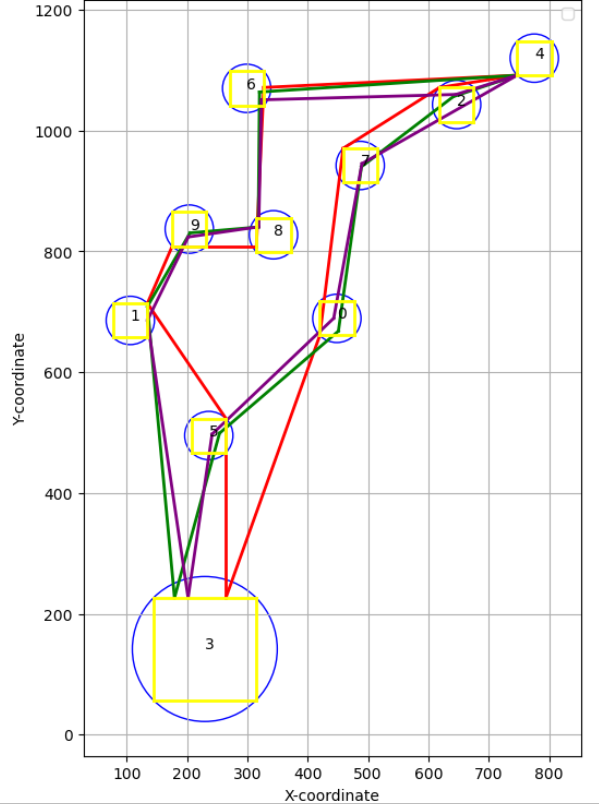}
     \caption{Trajectories for the instance ns10nb1.}
     \label{AbsL1L2Ns10Nb1}
\end{figure}

\begin{figure}[ht]
\centering
     \includegraphics[scale=0.25]{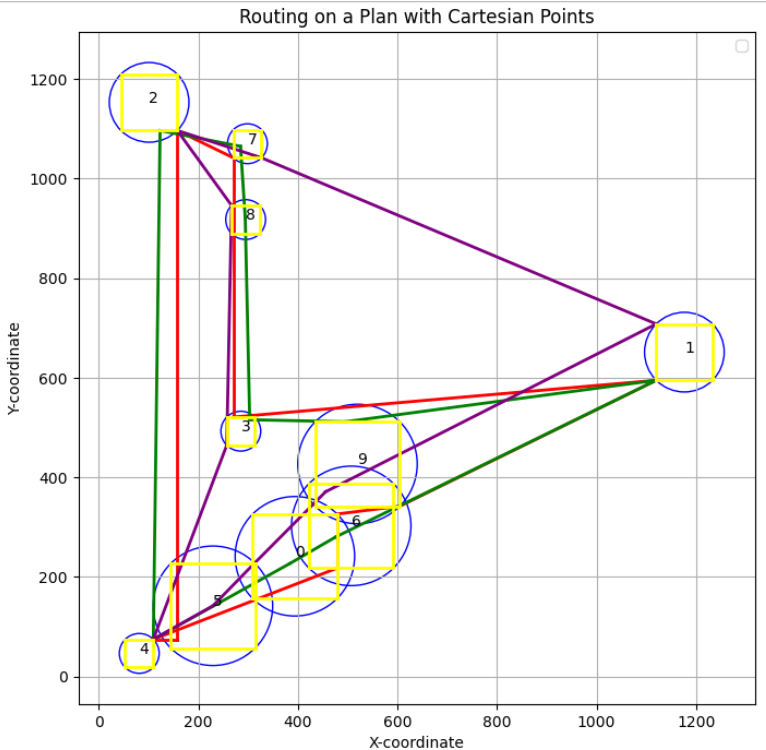}
     \caption{Trajectories for the instance ns10nb3.}
     \label{AbsL1L2Ns10Nb3}
\end{figure}

Some improved instances are illustrated by Figures \ref{AbsL1L2Ns10Nb1} and \ref{AbsL1L2Ns10Nb3}. The latter shows the trajectories found, considering small squares, for the instances "ns10nb1" and "ns10nb3" respectively. The red trajectories represent the solutions resulting from absolute value optimization. While, the green and purple trajectories are obtained from the linearized version of the model; namely linearization 1, and linearization 2 respectively. From these figures, we can see that the routings, resulting from the two linearizations, are now capable of also passing inside the considered convex set borders which was not the case with the absolute values.

\subsection{Testing linear regression}
The same set of instances are considered in order to compare the results of linear regression with those obtained by linearization 2.
 
Tables \ref{Lin2RegObjSSCPLEX} and \ref{Lin2RegObjSSMF} display the results for both small squares, obtained using the CPLEX solver directly and the Fragmented CPLEX-based Method, respectively. Furthermore, Tables \ref{Lin2RegObjSHCPLEX} and \ref{Lin2RegObjSHMF} showcase the results for both small hexagons, utilizing the CPLEX solver directly and the Fragmented CPLEX-based Method, respectively. For each instance presented in first column, tests with and without linear regression were carried out and are presented in the second and third multicolumns. For each, Linearization 2 is considered as it confirmed its efficiency. The complete solution is displayed in each multicolumn composed of the Manhattan cost, the Euclidian cost and the calculation times taken. the Man Euclidean distances found and the calculation times taken.

\begin{table}[!htb] \tiny
\caption{\scriptsize{Comparison of Linearization 2 and regression considering small square (CPLEX)}}
\begin{center}
    \begin{tabular}{|c|c|c|c|c|c|c|c|c|}
    \hline
   & \multicolumn{3}{c|}{Without Linear Regression} & \multicolumn{3}{c|}{With Linear Regression}  \\
  Instances & Manhattan  & Euclidean (m) & T (s) & Manhattan & Euclidean (m) & T (s)\\
  \hline
ns10nb0 & 3836.13 & 3086.33 &   12 & 3203.36 & \textcolor{red}{2599.68} & 18 \\ 
ns10nb1 & 2954.71 & 2482.45 & 21 & 2954.71 & \textcolor{red}{2439.51} & 20  \\ 
ns10nb2 & 5021.17 & 3979.06 &   7 & 4736.14 & \textcolor{red}{3843.9} & 6 \\ 
ns10nb3 & 4212.34 &  \textcolor{red}{3700.55} &   20 & 5528.6 & 4535.62 & 20 \\ 
ns10nb4 & 3030.57 &  \textcolor{red}{2347.23} &  20 & 3030.57 & 2390.71 & 21 \\ 
ns15nb0 & 3007.9 & \textcolor{red}{2548.96} & 16 & 3495.35 & 2918.14 & 6 \\ 
ns15nb1 & 3044.91 & \textcolor{red}{2428.83} & 16 & 3044.91 & 2458.7 & 18 \\ 
ns15nb2 & 3218.15 &  2725.63 & 21 & 3196.59 & \textcolor{red}{2674.71} & 21 \\ 
ns15nb3 & 5147.73 & 4343.03 & 22 & 5239.35 & \textcolor{red}{4313.79} & 19 \\ 
ns15nb4 & 4411.27 & 3825.95 & 18  & 4411.27 & \textcolor{red}{3660.23} & 18\\ 
ns20nb0 & 3536.67 & \textcolor{red}{2739.71} & 17 & 3523.19 & 2858.77 & 17 \\ 
ns20nb1 & 4183.69 & \textcolor{red}{3308.84} & 16 & 4183.68 & 3312.94 & 17 \\ 
ns20nb2 & 4529.3 & \textcolor{red}{3698} &18 & 4507.74 & 3766.5 & 17 \\ 
ns20nb3 & 5246.94 &  4609.92 & 18 & 5246.94 & \textcolor{red}{4595.66} & 17 \\ 
ns20nb4 & 4554.99 & 3931.43 & 16 & 4335.23 & \textcolor{red}{3844.15} & 17 \\ 
\hline
\end{tabular}
\end{center}
\label{Lin2RegObjSSCPLEX} 
\end{table}

\begin{table}[!htb] \tiny
\caption{\scriptsize{Comparison of Linearization 2 and regression considering small hexagons (CPLEX)}}
\begin{center}
    \begin{tabular}{|c|c|c|c|c|c|c|c|c|}
    \hline
  & \multicolumn{3}{c|}{Without Linear Regression} & \multicolumn{3}{c|}{With Linear Regression}  \\
  Instances & Manhattan  & Euclidean (m) & T (s) & Manhattan & Euclidean (m) & T (s)\\
  \hline
ns10nb0 & 3147.64 &  \textcolor{red}{2640.67} &  19  & 3901.56 &  3047.92 &  12 \\ 
ns10nb1 & 2896.99 &  \textcolor{red}{2449.77} &  24  & 2896.99 &  2450.75 &  21\\ 
ns10nb2 & 5223.6 &  4392.08 &  1  & 4724.42 &  \textcolor{red}{4048.86} &  3 \\ 
ns10nb3 & 4119.85 &  \textcolor{red}{3457.82} &  20  & 5723.06 &  4643.12 &  5\\ 
ns10nb4 & 2969.85 &  \textcolor{red}{2275.78} &  24  & 2969.85 &  2314.44 &  25\\ 
ns15nb0 & 2831.03 &  2468.08 &  19  & 2831.03 &  \textcolor{red}{2447.36} &  21\\ 
ns15nb1 & 2928.65 &  \textcolor{red}{2348.02} &  19  & 2928.65 &  2355.4 &  20\\ 
ns15nb2 & 3103.38 &  2729.25 &  21  & 3103.38 &  \textcolor{red}{2689.59} &  23\\ 
ns15nb3 & 5085.06 & 4318.33 &  23  & 5085.06 &  \textcolor{red}{4310.33} &  23\\ 
ns15nb4 & 4361.74 & 3708.76 &  20  & 4338.86 &  \textcolor{red}{3624.43} &  22 \\ 
ns20nb0 & 3272.17 &  \textcolor{red}{2701.82} &  18  & 3268.17 &  2816.34 &  17 \\ 
ns20nb1 & 4007.83 &  \textcolor{red}{3157.06} &  19  & 4007.86 &  3326.27 &  18 \\ 
ns20nb2 & 4244.01 & 3634.18 &  18  & 4244.13 &  \textcolor{red}{3612.97} &  19\\ 
ns20nb3 & 5050.3 &  \textcolor{red}{4479.34} &  19  & 5050.3 &  4483.6 &  17 \\ 
ns20nb4 & 4260.63 &  \textcolor{red}{3733.97} &  23  & 4260.63 &  3787.06 &  18\\ 
\hline
\end{tabular}
\end{center}
\label{Lin2RegObjSHCPLEX} 
\end{table}

\begin{table}[!htb] \tiny
\caption{\scriptsize{Comparison of Linearization 2 and regression considering small square (MF)}}
\begin{center}
    \begin{tabular}{|c|c|c|c|c|c|c|c|c|}
    \hline
   & \multicolumn{3}{c|}{Without Linear Regression} & \multicolumn{3}{c|}{With Linear Regression}  \\
  Instances & Manhattan  & Euclidean (m) & T (s) & Manhattan & Euclidean (m) & T (s)\\
  \hline
ns10nb0.txt & 3200.05 & 2552.47 & 0 &  3199.65 & 2552.46 & 0 \\ 
ns10nb1.txt & 3090.16 & 2413.92 & 0 &  3090.16 & 2413.92 & 0 \\ 
ns10nb2.txt & 3978.83 & 3146.29 & 0 &  3978.83 & 3146.29 & 0 \\ 
ns10nb3.txt & 4282.78 & 3340.61 & 0 &  4282.78 & 3340.61 & 0 \\ 
ns10nb4.txt & 3129.99 & 2330.85 & 0 &  3129.99 & 2330.85 & 0 \\ 
ns15nb0.txt & 3120.76 & 2441 & 0 &  3120.76 & 2441 & 0 \\ 
ns15nb1.txt & 3032.04 & 2351.23 & 0 &  3032.04 & 2351.23 & 0 \\ 
ns15nb2.txt & 3205.49 & 2607.84 & 0 &  3205.49 & 2607.84 & 0 \\ 
ns15nb3.txt & 5661.2 & 4286.27 & 0 &  5661.2 & 4286.27 & 0 \\ 
ns15nb4.txt & 4557.01 & 3508.48 & 0 &  4557.01 & 3508.48 & 0 \\ 
ns20nb0.txt & 3568.99 & \textcolor{red}{2673.4} & 0 &  3568.99 & 2676.4 & 0 \\ 
ns20nb1.txt & 4169.3 & 3256.11 & 0 &  4169.3 & 3256.11 & 0 \\ 
ns20nb2.txt & 4636.26 & 3725.22 & 2 &  4636.26 & 3725.22 & 2 \\ 
ns20nb3.txt & 5426.52 & 4257.57 & 0 &  5446.19 & \textcolor{red}{4250.76} & 0 \\ 
ns20nb4.txt & 4359.87 & 3671.67 & 0 &  4359.87 & 3671.67 & 0 \\ 
ns50nb0.txt & 6255.9 & 5066.96 & 2 &  6255.41 & \textcolor{red}{5061.76} & 2 \\ 
ns50nb1.txt & 6589.43 & \textcolor{red}{5159.29} & 19 &  6624.37 & 5192.86 & 18 \\ 
ns50nb2.txt & 5897.93 & 4764.3 & 1 &  5911.94 & \textcolor{red}{4718.42} & 2 \\ 
ns50nb3.txt & 6021.42 & 4797.71 & 2 &  6021.42 & 4797.71 & 2 \\ 
ns50nb4.txt & 7258.89 & \textcolor{red}{5524.44} & 7 &  7302.66 & 5540.47 & 7 \\ 
ns70nb0.txt & 7793.16 & \textcolor{red}{6155.21} & 1059 &  7805.86 & 6160.75 & 437 \\  
\hline
\end{tabular}
\end{center}
\label{Lin2RegObjSSMF} 
\end{table}

\begin{table}[!htb] \tiny
\caption{\scriptsize{Comparison of Linearization 2 and regression considering small hexagons (MF)}}
\begin{center}
    \begin{tabular}{|c|c|c|c|c|c|c|c|c|}
    \hline
  & \multicolumn{3}{c|}{Without Linear Regression} & \multicolumn{3}{c|}{With Linear Regression}  \\
  Instances & Manhattan  & Euclidean (m) & T (s) & Manhattan & Euclidean (m) & T (s)\\
  \hline
ns10nb0.txt & 3242.96 & 2568.88 & 0 &  3242.96 & 2568.88 & 0 \\ 
ns10nb1.txt & 2947.01 & 2370.38 & 0 &  2947.01 & 2370.38 & 0 \\ 
ns10nb2.txt & 3947.71 & \textcolor{red}{3119.68} & 0 &  3939.67 & 3166.98 & 0 \\ 
ns10nb3.txt & 4224.05 & 3291.44 & 0 &  4224.05 & \textcolor{red}{3281.94} & 0 \\ 
ns10nb4.txt & 3057.7 & 2290.83 & 0 &  3057.7 & 2290.83 & 0 \\ 
ns15nb0.txt & 3034.75 & \textcolor{red}{2439.94} & 0 &  3034.75 & 2441.51 & 0 \\ 
ns15nb1.txt & 2970.47 & 2300.3 & 0 &  2970.47 & 2300.3 & 0 \\ 
ns15nb2.txt & 3125.26 & 2553.88 & 0 &  3125.26 & 2553.88 & 0 \\ 
ns15nb3.txt & 5565.47 & 4210.51 & 0 &  5565.47 & 4210.51 & 0 \\ 
ns15nb4.txt & 4428.42 & 3499.17 & 0 &  4428.42 & 3499.17 & 0 \\ 
ns20nb0.txt & 3411.64 & 2587.97 & 0 &  3411.64 & 2587.97 & 0 \\ 
ns20nb1.txt & 4087.68 & 3183.05 & 0 &  4087.68 & 3183.05 & 0 \\ 
ns20nb2.txt & 4389.47 & 3611.85 & 2 &  4389.47 & 3611.85 & 2 \\ 
ns20nb3.txt & 5255.3 & 4216.64 & 0 &  5255.3 & 4216.55 & 0 \\ 
ns20nb4.txt & 4318.33 & \textcolor{red}{3652.86} & 0 &  4480.33 & 3741.39 & 1 \\ 
ns50nb0.txt & 6200.96 & 5142.52 & 3 &  6200.96 & \textcolor{red}{5138.69} & 1 \\ 
ns50nb1.txt & 6336.82 & \textcolor{red}{5038.75} & 23 &  6406.43 & 5095.6 & 20 \\ 
ns50nb2.txt & 5717.04 & \textcolor{red}{4655.77} & 2 &  5699.55 & 4658.58 & 2 \\ 
ns50nb3.txt & 5924.36 & \textcolor{red}{4764.53} & 1 &  5924.36 & 4773.91 & 2 \\ 
ns50nb4.txt & 7123.6 & \textcolor{red}{5432.24} & 6 &  7125.4 & 5447.73 & 6 \\ 
ns70nb0.txt & 7737.19 & \textcolor{red}{6116.44} & 228 &  7760.01 & 6175.74 & 155 \\ 
\hline
\end{tabular}
\end{center}
\label{Lin2RegObjSHMF} 
\end{table}  

Tables \ref{Lin2RegObjSSCPLEX} and \ref{Lin2RegObjSHCPLEX} reveal a competitive trend for both formulations, with and without linear regression, when utilizing the CPLEX solver on the complete model. Specifically, a $53.33\%$ improvement is observed when considering the convex set of small squares, and a $40\%$ improvement is observed when considering small hexagons.

Conversely, Tables \ref{Lin2RegObjSSMF} and \ref{Lin2RegObjSHMF} indicate fewer improvements when employing the Fragmented CPLEX-based Method. Specifically, out of twenty-one solutions for small squares, only three showed improvement, while the same costs were obtained for 14 instances, accounting for approximately $80.9\%$ of similar or better performance. Additionally, two solutions out of twenty-one were enhanced for small hexagons, and a similar performance was observed for eleven instances, representing around $61.9\%$ of better or equal costs.

Additionally, it is observed from Tables \ref{Lin2RegObjSSMF} and \ref{Lin2RegObjSHMF} that the majority of solutions where Linear Regression failed to bring improvement were those with a larger number of sensors.

\subsection{Testing 8D projection}
The same set of instances are considered in order to compare the results of applying 8D projection in addition to Linearized objective function using Linearization 2 and linear regression. 

Table \ref{Lin2RegProjObjSHMF} shows the results considering small hexagons, respectively. For each instance presented in first column, tests without and with 8D projection were carried out and are presented in the second and third multicolumns respectively for each tested instance indicated in first column. 
The complete solution is displayed in each multicolumn composed of the Manhattan cost, the Euclidian cost and the calculation times taken. 

\begin{table}[!htb] \tiny
\caption{\scriptsize{Comparison of 8D projection considering small hexagon (MF)}}
\begin{center}
    \begin{tabular}{|c|c|c|c|c|c|c|c|c|}
    \hline
   & \multicolumn{3}{c|}{Without 8D projection} & \multicolumn{3}{c|}{With 8D Projection}  \\
  Instances & Manhattan  & Euclidean (m) & T (s) & Manhattan & Euclidean (m) & T (s)\\
  \hline
ns10nb0.txt & 3242.96 & 2568.88 & 0 &  3190.8 & \textcolor{red}{2531} & 0 \\ 
ns10nb1.txt & 2947.01 & 2370.38 & 0 &  2947.01 & \textcolor{red}{2364.63} & 0 \\ 
ns10nb2.txt & 3939.67 & 3166.98 & 0 &  3947.71 & \textcolor{red}{3111.3} & 0 \\ 
ns10nb3.txt & 4224.05 & \textcolor{red}{3281.94} & 0 &  4224.05 & 3283.32 & 0 \\ 
ns10nb4.txt & 3057.7 & 2290.83 & 0 &  3057.7 & \textcolor{red}{2288.57} & 0 \\ 
ns15nb0.txt & 3034.75 & 2441.51 & 0 &  2997.73 & \textcolor{red}{2405.62} & 0 \\ 
ns15nb1.txt & 2970.47 & 2300.3 & 0 &  2988.83 & \textcolor{red}{2294.85} & 0 \\ 
ns15nb2.txt & 3125.26 & \textcolor{red}{2553.88} & 0 &  3125.26 & 2576.66 & 0 \\ 
ns15nb3.txt & 5565.47 & 4210.51 & 0 &  5565.47 & \textcolor{red}{4200.33} & 0 \\ 
ns15nb4.txt & 4428.42 & \textcolor{red}{3499.17} & 0 &  4442.94 & 3502.32 & 0 \\ 
ns20nb0.txt & 3411.64 & 2587.97 & 0 &  3411.64 & \textcolor{red}{2585.33} & 0 \\ 
ns20nb1.txt & 4087.68 & 3183.05 & 0 &  4034.4 & \textcolor{red}{3162.7} & 0 \\ 
ns20nb2.txt & 4389.47 & 3611.85 & 2 &  4348.19 & \textcolor{red}{3562.42} & 2 \\ 
ns20nb3.txt & 5255.3 & \textcolor{red}{4216.55} & 0 &  5255.3 & 4220.46 & 0 \\ 
ns20nb4.txt & 4480.33 & 3741.39 & 1 &  4333.91 & \textcolor{red}{3658.85} & 1 \\ 
ns50nb0.txt & 6200.96 & 5138.69 & 1 &  6160.8 & \textcolor{red}{5085.7} & 2 \\ 
ns50nb1.txt & 6406.43 & 5095.6 & 20 &  6307.25 & \textcolor{red}{5047.58} & 48 \\ 
ns50nb2.txt & 5699.55 & \textcolor{red}{4658.58} & 2 &  5691.04 & 4672.57 & 1 \\ 
ns50nb3.txt & 5924.36 & 4773.91 & 2 &  5923.6 & \textcolor{red}{4738.78} & 2 \\ 
ns50nb4.txt & 7125.4 & 5447.73 & 6 &  7103.84 & \textcolor{red}{5416.65} & 6 \\ 
ns70nb0.txt & 7760.01 & \textcolor{red}{6175.74} & 155 &  7805.59 & 6283.11 & 122 \\ 
\hline
\end{tabular}
\end{center}
\label{Lin2RegProjObjSHMF} 
\end{table}

From the data presented in Table \ref{Lin2RegProjObjSHMF}, it is evident that there is a notable improvement in the displacement cost for approximately $71.428\%$ of the tested instances, as indicated by the instances highlighted in red. On average, the solutions were enhanced by approximately $31.68\%$. It is noteworthy that the effectiveness of the 8D projection does not appear to be dependent on the size of the instances.

Moreover, it is important to highlight that the integration of the 8D projection did not lead to an increase in computation time. This is because the 8D projection is solely incorporated into the objective function and does not entail a separate computational process.
\subsection{Discussion}

To draw clear conclusions about which type of linearization or convex set shape has provided most of the lowest costs, the Table \ref{ComparAll} is displayed to summarize the best solutions obtained so far (in terms of the Euclidean metric). The first column presenting the names of the instances is followed by a multicolumn of size three presenting the best solution found so far on the different formulations and configurations. The latter shows the Manhattan, Euclidean distances, and the calculation time. Then, the third column indicates which convex set shape was part of the mathematical formulation (PC: small squares, PH: small hexagons). The fourth column shows  which of the two linearizations made it possible to obtain the displayed solution, if it exists, linearization 1 (Lin1), linearization 2 (Lin2), or Absolute value (ABS) otherwise. The fifth column precises if the linear regression (Reg) coefficients is part of some configurations that helped to find the solution or not, or both cases (Reg/). In the same manner, the sixth column indicates if 8D Projection was used (Proj) or not, or both cases (Proj/). Finally, the last column indicates which solution method was used; CPLEX or fragmented CPLEX-based method.

\begin{table}[!htb] \tiny
\caption{\scriptsize{Best solutions found among all mathematical formulations}}
\begin{center}
    \begin{tabular}{|c|c|c|c|c|c|c|c|c|c|}
    \hline
  Instances & \multicolumn{3}{c|}{Best cost} & Best  & Best & & & Best   \\
   & Manhattan  & Euclidean (m) & T (s)  & linéarization & convex set & Linear Regression & Projection & method \\
  \hline
ns10nb0 & 3190.8 & 2530.74 & 0 & PH & ABS & Reg/ & Proj & MF \\
ns10nb1 & 2947.01 & 2359.83 & 1 & PH & ABS & Reg &  & MF \\
ns10nb2 & 3939.67 & 3105.74 & 0 & PH & ABS & Reg/ & Proj & MF \\
ns10nb3 & 4224.05 & 3281.94 & 0 & PH & Lin2 & Reg &  & MF  \\
ns10nb4 & 3057.7 & 2288.57 & 0 & PH & Lin2 & Reg & Proj  & MF  \\
ns15nb0 & 2997.73 & 2405.62 & 0 & PH & Lin2 & Reg & Proj & MF  \\
ns15nb1 & 2988.83 & 2294.85 & 0 & ,PH & Lin2 & Reg & Proj  & MF \\
ns15nb2 & 3125.26 & 2553.88 & 0 & PH & Lin2 & Reg/ &  & MF \\
ns15nb3 & 5565.47 & 4199.56 & 0 & ,PH & Lin2 & & Proj  & MF \\
ns15nb4 & 4468.42 & 3495.34 & 0 & PH & ABS & Reg/ & Proj/ & MF \\
ns20nb0 & 3411.64 & 2585.33 & 0 & PH & Lin2 & Reg & Proj  & MF \\
ns20nb1 & 4031.99 & 3145.46 & 2 & PH & ABS & Reg/ & Proj & MF \\
ns20nb2 & 4398.19 & 3512.68 & 6 & PH & ABS & Reg & Proj  & MF \\
ns20nb3 & 5255.3 & 4216.55 & 0 & PH & Lin2 & Reg &  & MF \\
ns20nb4 & 4318.33 & 3652.86 & 0 & PH & Lin2 & & Proj/ & MF \\
ns50nb0 & 6122.39 & 5016.17 & 9 & PH & ABS & &  & MF \\
ns50nb1 & 6336.82 & 5038.75 & 23 & ,PH & Lin2 & &  & MF \\
ns50nb2 & 5781.73 & 4641.79 & 16 & PH & ABS & & Proj  & MF \\
ns50nb3 & 5923.6 & 4738.78 & 2 & PH & Lin2 & Reg & Proj  & MF \\
ns50nb4 & 7100.68 & 5415.09 & 10 & PH & ABS & Reg &  & MF \\
ns70nb0 & 7737.19 & 6058.92 & 202 & PH & Lin2 & & Proj & MF \\
  \hline
\end{tabular}
\end{center}
\label{ComparAll} 
\end{table}

From this last Table \ref{ComparAll}, we notice that small hexagons (PH), especially when used in a fragmented method (MF), provide all of the best achieved solutions, in terms of travel cost (Euclidean). 
Furthermore, the fragmented method seems to improve both the travel cost and the CPU calculation time, compared to CPLEX. Moreover, $57.14\%$ of these solutions were found after applying linearization 2 and the other $38.1\%$ of them are reached without any linearization. Only $3$ best solutions were found using absolute value formulation. Despite the fact that linear regression didn't seem improving the solutions found by the absolute value or Linearization 2 formulations, it was able to reach $71.43\%$ of the best solutions, where $47.6\%$ of them were obtained using 8D projection at the same time. At the same time, a lower proportion of $52.4\%$ from these solutions were found without linear regression. Finally, $66.6\%$ of the best solutions are achieved using the 8D projection and only  $42.8\%$ were found without it.

\begin{figure}[ht]
\centering
     \includegraphics[scale=0.30]{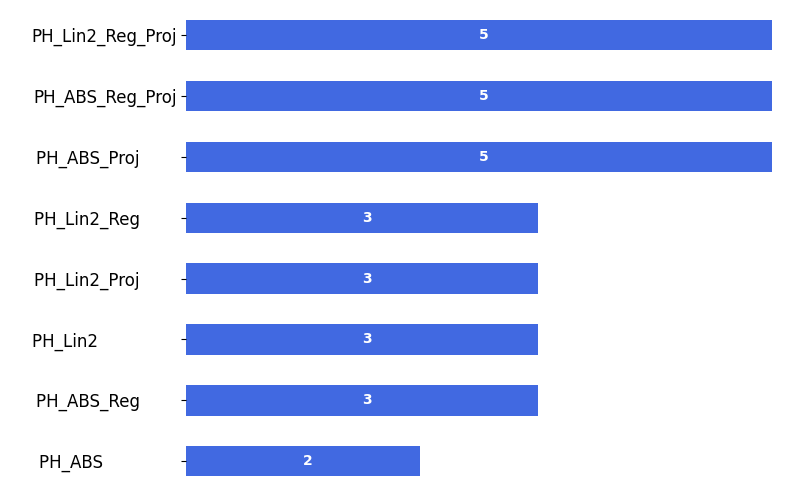}
     \caption{The Number of best reached solutions for each configuration}
     \label{NBestSolutions}
\end{figure}

This last table shows the contribution of each formulations into the process of optimization of the overall problem. However, Figure \ref{NBestSolutions} presents a bar chart in order to show the configurations that were able to achieve the solution reported in Table \ref{ComparAll} and the length of the bar is relative to the number of best solution reached by each one. The bar are ordered in decreasing order from top to bottom. Clearly, the two configurations related to the longer bars are "PH-Lin2-Reg-Proj" and "PH-ABS-Reg-Proj" even if there is not a big difference between them. 

\begin{figure}[ht]
\centering
     \includegraphics[scale=0.30]{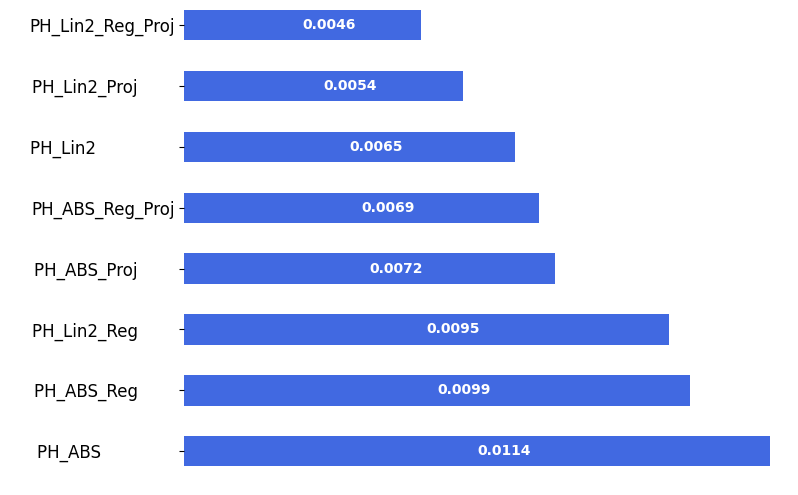}
     \caption{The average relative error of all instances}
     \label{RelativeErrorAvg}
\end{figure}

Moreover, the relative error $RE$ of the obtained displacement cost $AC$ regarding the best one reached so far $BC$ is calculated for each instance, for each configuration. The formula used is as follows : $RE = \frac{(AC - BC)}{BC}$. Then, the average relative error $ARE$ is calculated overall the tested instances for all formulations. Therefore, $ARE = \frac{RE}{21}$ as there are 21 instances tested. The Figure \ref{RelativeErrorAvg} shows the average relative error for each formulation using a bar chart. The bars are sorted in a decreasing order to go from the formulation that have the lower error to the one that have the larger error. 

From this figure, it is noticed that the average errors are quite close to each other. However, the formulations with linearization 2, with or without projection, have the lower average relative error. Then, the formulations considering absolute values in the objective function comes after with a bigger absolute values. Once again, the linear regression, when used without projection, is worsening the average relative error for both absolute value and linearization 2. However, this latter is enhancing the the average relative error and the number of reached best costs for both formulations. Finally, "PH-ABS"comes at the end of the plot as it has the worst average relative error. 

\section{Conclusions}
Throughout this study, the Close Enough Traveling Salesman Problem (CETSP) has emerged as a critical challenge in optimizing unmanned aerial vehicle (UAV) trajectory planning. The CETSP, which relaxes the strict requirement of visiting every point exactly once in the classic Traveling Salesman Problem (TSP), mirrors real-world scenarios where close proximity is acceptable for visiting locations. In the realm of UAV trajectory planning, where efficiency, resource utilization, and path optimization are paramount, addressing the CETSP is imperative to ensure the practicality and effectiveness of our proposed approach. 

To overcome the complexities posed by the CETSP, we introduce two simplification strategies: linearization and approximation of the Euclidean cost. We explore two linearization options and compare their effectiveness. Additionally, we employ a linear regression machine learning technique to derive coefficients that map Manhattan distances to Euclidean distances. Further enhancing the approximation, we investigate eight projections of the displacement cost on new virtual axes within the optimized function.

The two linearizations manage to guide the drone through the entire convex space. However, linearization 2 manages to optimize the trajectory cost more effectively for the overall tested cases. Moreover,
The combination between the fragmented method and small hexagons made it possible to achieve the best travel costs for the majority of instances, on both small and large instances. Furthermore, the implementation of the 8D projection significantly bolstered the effectiveness of reducing displacement costs.

Although linear regression, when used with the 8D Projection at the same time, shows promising prospects, it is possible to improve its results by integrating other key parameters into the learning due to the fact that the hitting points coordinates are different than the center ones and are not known in advance.

Cognizant of our findings, future research will explore improving the coefficients from linear regression by adding other key parameters to learning, and adapting our approach to accommodate obstacles within the map, ensuring its relevance across diverse use cases.

\bibliographystyle{elsarticle-num} 
\bibliography{cas-refs}

\begin{thebibliography}{10}
\expandafter\ifx\csname url\endcsname\relax
  \def\url#1{\texttt{#1}}\fi
\expandafter\ifx\csname urlprefix\endcsname\relax\def\urlprefix{URL }\fi
\expandafter\ifx\csname href\endcsname\relax
  \def\href#1#2{#2} \def\path#1{#1}\fi

\bibitem{cariou2023evolutionary}
C.~Cariou, L.~Moiroux-Arvis, F.~Pinet, J.-P. Chanet, Evolutionary algorithm with geometrical heuristics for solving the close enough traveling salesman problem: Application to the trajectory planning of an unmanned aerial vehicle, Algorithms 16~(1) (2023) 44.

\bibitem{hou2024natural}
Y.~Hou, X.~Wang, H.~Chang, Y.~Dong, D.~Zhang, C.~Wei, I.~Lee, Y.~Yang, Y.~Liu, J.~Zhang, Natural gas consumption monitoring based on k-nn algorithm and consumption prediction framework based on backpropagation neural network, Buildings 14~(3) (2024) 627.

\bibitem{patelradio}
K.~Patel, M.~Kandasamy, R.~Shanmugam, T.~Uttare, N.~Samani, M.~Aqeel, S.~Joshi, Radio frequency identification and authentication-based intelligent parking management system using the iot and mobile applications: An implementation point of view, in: Green Computing for Sustainable Smart Cities, CRC Press, pp. 138--149.

\bibitem{diaz2024mathematical}
D.~D{\'\i}az-R{\'\i}os, J.-J. Salazar-Gonz{\'a}lez, Mathematical formulations for consistent travelling salesman problems, European Journal of Operational Research 313~(2) (2024) 465--477.

\bibitem{csenturk2020steiner}
{\.I}.~{\c{S}}enturk, A steiner zone approach for mobile data collection in partitioned wireless sensor networks, Bili{\c{s}}im Teknolojileri Dergisi 13~(3) (2020) 217--224.

\bibitem{Korte2008}
\href{https://doi.org/10.1007/978-3-540-71844-4_21}{The Traveling Salesman Problem}, Springer Berlin Heidelberg, Berlin, Heidelberg, 2008, pp. 527--562.
\newblock \href {https://doi.org/10.1007/978-3-540-71844-4_21} {\path{doi:10.1007/978-3-540-71844-4_21}}.
\newline\urlprefix\url{https://doi.org/10.1007/978-3-540-71844-4_21}

\bibitem{mennell2009heuristics}
W.~K. Mennell, Heuristics for solving three routing problems: Close-enough traveling salesman problem, Close-Enough Vehicle Routing Problem, and sequence-dependent team orienteering problem, University of Maryland, College Park, 2009.

\bibitem{pacheco2023exponential}
T.~Pacheco, R.~Martinelli, A.~Subramanian, T.~A. Toffolo, T.~Vidal, Exponential-size neighborhoods for the pickup-and-delivery traveling salesman problem, Transportation Science 57~(2) (2023) 463--481.

\bibitem{pop2023comprehensive}
P.~C. Pop, O.~Cosma, C.~Sabo, C.~P. Sitar, A comprehensive survey on the generalized traveling salesman problem, European Journal of Operational Research (2023).

\bibitem{ota2023flow}
C.~T. Ota, D.~J. Fiorotto, C.~T. L. d.~S. Ghidini, W.~A.~d. Oliveira, A flow-based model for the multivehicle covering tour problem with route balancing, International Transactions in Operational Research (2023).

\bibitem{gulczynski2006close}
D.~J. Gulczynski, J.~W. Heath, C.~C. Price, The close enough traveling salesman problem: A discussion of several heuristics, Perspectives in Operations Research: Papers in Honor of Saul Gass’ 80 th Birthday (2006) 271--283.

\bibitem{yuan2007optimal}
B.~Yuan, M.~Orlowska, S.~Sadiq, On the optimal robot routing problem in wireless sensor networks, IEEE transactions on knowledge and data engineering 19~(9) (2007) 1252--1261.

\bibitem{mennell2011steiner}
W.~Mennell, B.~Golden, E.~Wasil, A steiner-zone heuristic for solving the close-enough traveling salesman problem, in: 2th INFORMS computing society conference: operations research, computing, and homeland defense, 2011.

\bibitem{behdani2014integer}
B.~Behdani, J.~C. Smith, An integer-programming-based approach to the close-enough traveling salesman problem, INFORMS Journal on Computing 26~(3) (2014) 415--432.

\bibitem{coutinho2016branch}
W.~P. Coutinho, R.~Q.~d. Nascimento, A.~A. Pessoa, A.~Subramanian, A branch-and-bound algorithm for the close-enough traveling salesman problem, INFORMS Journal on Computing 28~(4) (2016) 752--765.

\bibitem{carrabs2017novel}
F.~Carrabs, C.~Cerrone, R.~Cerulli, M.~Gaudioso, A novel discretization scheme for the close enough traveling salesman problem, Computers \& Operations Research 78 (2017) 163--171.

\bibitem{yang2018double}
Z.~Yang, M.-Q. Xiao, Y.-W. Ge, D.-L. Feng, L.~Zhang, H.-F. Song, X.-L. Tang, A double-loop hybrid algorithm for the traveling salesman problem with arbitrary neighbourhoods, European Journal of Operational Research 265~(1) (2018) 65--80.

\bibitem{wang2019steiner}
X.~Wang, B.~Golden, E.~Wasil, A steiner zone variable neighborhood search heuristic for the close-enough traveling salesman problem, Computers \& Operations Research 101 (2019) 200--219.

\bibitem{semami2019close}
S.~Semami, H.~Toulni, A.~Elbyed, The close enough traveling salesman problem with time window, Int. J. Circuits Syst. Signal Process 13 (2019) 579--584.

\bibitem{fanta2021close}
L.~Fanta, The close enough travelling salesman problem in the polygonal domain, Ph.D. thesis, Master’s thesis, CTU in Prague (2021).

\bibitem{di2023generalized}
A.~Di~Placido, C.~Archetti, C.~Cerrone, B.~Golden, The generalized close enough traveling salesman problem, European Journal of Operational Research 310~(3) (2023) 974--991.

\bibitem{qian2023solving}
Q.~Qian, Y.~Wang, D.~Boyle, On solving close enough orienteering problem with overlapped neighborhoods, arXiv preprint arXiv:2310.04257 (2023).

\bibitem{sinha2021estimating}
D.~Sinha~Roy, B.~Golden, X.~Wang, E.~Wasil, Estimating the tour length for the close enough traveling salesman problem, Algorithms 14~(4) (2021) 123.

\bibitem{camino2019linearization}
J.-T. Camino, C.~Artigues, L.~Houssin, S.~Mourgues, Linearization of euclidean norm dependent inequalities applied to multibeam satellites design, Computational Optimization and Applications 73 (2019) 679--705.

\end{thebibliography}

\end{document}